\documentclass{article}
\usepackage[final]{arxiv_nips_2017}

\usepackage[utf8]{inputenc} 
\usepackage[T1]{fontenc}    
\usepackage{hyperref}       
\usepackage{url}            
\usepackage{booktabs}       
\usepackage{microtype}      
\usepackage{nicefrac}       
\usepackage{authblk}
\usepackage{enumitem}

\usepackage{amsmath,amssymb,amsfonts}
\usepackage{amsthm}

\newcommand{\ts}{\textstyle}
\newcommand{\lossyxh}{\text{loss}(y_d, \hat{y}(x_d, h_d, \xi^y))}

\usepackage{parskip}

\usepackage{hyperref}
\hypersetup{
    colorlinks,
    citecolor=black,
    filecolor=black,
    linkcolor=black,
    urlcolor=black,
}

\usepackage{color}

\usepackage{natbib}
\setcitestyle{authoryear,open={(},close={)}}

\usepackage{graphicx}
\graphicspath{{figures/}}

\usepackage[font=small]{caption}
\usepackage{float}
\usepackage{algorithm}
\usepackage[noend]{algpseudocode}
\algrenewcommand\algorithmicfunction{\textbf{def}}
\algrenewcommand\algorithmicrequire{\textbf{Input:}}
\algrenewcommand\algorithmicensure{\textbf{Output:}}

\makeatletter
\newcommand{\verbatimfont}[1]{\renewcommand{\verbatim@font}{\ttfamily#1}}
\makeatother

\title{Prediction-Constrained Training for\\ Semi-Supervised Mixture and Topic Models}

\usepackage{authblk}
\author[1]{\textbf{Michael C. Hughes}\thanks{Contact email:~\texttt{mike@michaelchughes.com}}~~}
\author[2]{\textbf{Leah Weiner}}
\author[3]{\textbf{Gabriel Hope}}
\author[4]{\textbf{Thomas H. McCoy, Jr.}}
\author[4]{\textbf{Roy H. Perlis}}
\author[3]{\textbf{Erik B. Sudderth}}
\author[1]{\textbf{Finale Doshi-Velez}}
\affil[1]{School of Engineering and Applied Sciences, Harvard University}
\affil[2]{Dept. of Computer Science, Brown University}
\affil[3]{School of Information and Computer Sciences, Univ. of California, Irvine}
\affil[4]{Massachusetts General Hospital}

\begin{document}

\setlength{\abovedisplayskip}{2pt plus 3pt}
\setlength{\belowdisplayskip}{2pt plus 3pt}

\maketitle 

\begin{abstract}
Supervisory signals have the potential to make low-dimensional data representations, like those learned by mixture and topic models, more interpretable and useful.  
We propose a framework for training latent variable models that explicitly balances two goals: recovery of faithful generative explanations of high-dimensional data, and accurate prediction of associated semantic labels.
Existing approaches fail to achieve these goals due to an incomplete treatment of a fundamental asymmetry: the intended application is always predicting labels from data, not data from labels.  
Our \emph{prediction-constrained} objective for training generative models coherently integrates loss-based supervisory signals while enabling effective semi-supervised learning from partially labeled data.
We derive learning algorithms for semi-supervised mixture and topic models using stochastic gradient descent with automatic differentiation.
We demonstrate improved prediction quality compared to several previous supervised topic models,
achieving predictions competitive with high-dimensional logistic regression on text sentiment analysis and electronic health records tasks while simultaneously learning interpretable topics.
\end{abstract}

\section{Introduction}

Latent variable models are widely used to explain high-dimensional data by learning appropriate low-dimensional structure.
For example, a model of online restaurant reviews
might describe a single user's long plain text as a blend of terms describing customer service and terms related to Italian cuisine.
When modeling electronic health records,
a single patient's high-dimensional medical history of lab results and diagnostic reports might be described as a classic instance of juvenile diabetes.
Crucially, we often wish to discover a faithful low-dimensional representation rather than rely on restrictive predefined representations.
\emph{Latent variable models} (LVMs), including
mixture models and topic models like Latent Dirichlet Allocation \citep{blei2003lda}, are widely used for \emph{unsupervised} learning from high-dimensional data.
There have been many efforts to generalize these methods to \emph{supervised} applications in which observations are accompanied by target values, especially when we seek to predict these targets from future examples.
For example, \citet{paul2012model} use topics from Twitter to model trends in flu, and \citet{jiang2015travel} use topics from image captions to make travel recommendations.  By smartly capturing the joint distribution of input data and targets, supervised LVMs may lead to predictions that better generalize from limited training data.  Unfortunately, many previous methods for the supervised learning of LVMs fail to deliver on this promise---in this work, our first contribution is to provide theoretical and empirical explanation that exposes fundamental problems in these prior formulations.

One na\"ive application of LVMs like topic models to supervised tasks uses \emph{two-stage} training: first train an unsupervised model, and then train a supervised predictor given the fixed latent representation from stage one.
Unfortunately, this two-stage pipeline often fails to produce high-quality predictions, especially when the raw data features are not carefully engineered and contain structure irrelevant for prediction. For example, applying LDA to clinical records might find topics about common conditions like diabetes or heart disease, which may be irrelevant if the ultimate supervised task is predicting sleep therapy outcomes.  

Because this two-stage approach is often unsatisfactory, many attempts have been made to directly incorporate supervised labels as observations in a single generative model.  
For mixture models, examples of supervised training are numerous~\citep{hannah2011DPmixGLM,shahbaba2009nonlinearDPmix}.
Similarly, many topic models have been proposed that jointly generate word counts and document labels~\citep{blei2007sLDA, lacoste2009disclda, wang2009simultaneous, zhu2012medlda, chen2015bplda}.  However, a survey by \citet{halpern2012comparison} finds that these approaches have little benefit, if any, over standard unsupervised LDA in clinical prediction tasks. Furthermore,  often the quality of supervised topic models does not significantly improve as model capacity (the number of topics) increases, even when large training datasets are available.

In this work, we expose and correct several deficiencies in previous formulations of supervised topic models.  We introduce a learning objective that directly enforces the intuitive goal of representing the data in a way that enables accurate downstream predictions.  Our objective acknowledges the inherent asymmetry of prediction tasks: a clinician is interested in predicting sleep outcomes given medical records, but not medical records given sleep outcomes.  Approaches like \emph{supervised LDA} (sLDA,~\citet{blei2007sLDA}) that optimize the joint likelihood of labels and words ignore this crucial asymmetry.  Our \emph{prediction-constrained} latent variable models are tuned to maximize the marginal likelihood of the observed data, subject to the constraint that prediction accuracy (formalized as the conditional probability of labels given data) exceeds some target threshold.

We emphasize that our approach seeks to find a compromise between two distinct goals: 
build a reasonable density model of observed data while making high-quality predictions of some target values given that data. 
If we only cared about modeling the data well, we could simply ignore the target values and adapt standard frequentist or Bayesian training objectives.
If we only cared about prediction performance, there are a host of discriminative regression and classification methods.
However, we find that many applications benefit from the representations which LVMs provide, 
including the ability to explain target predictions from high-dimensional data via an interpretable low-dimensional representation.  In many cases, introducing supervision enhances the interpretability of the generative model as well, as the task forces modeling effort to focus on only relevant parts of high-dimensional data.  Finally, in many applications it is beneficial to have the ability to learn from observed data for which target labels are unavailable.
We find that especially in this \emph{semi-supervised} domain, our prediction-constrained training objectives provides clear wins over existing methods.


\section{Prediction-constrained Training for Latent Variable Models}

In this section, we develop a prediction-constrained
training objective applicable to a broad family of latent variable models. Later
sections provide concrete learning algorithms for supervised variants of mixture models
\citep{everitt1981mixtures} and topic models \citep{blei2012topicmodels}. However, we emphasize
that this framework could be applied much more broadly to allow supervised
training of well-known 
generative models like probabilistic PCA \citep{roweis1998algorithms, tipping1999probabilisticPCA},
dynamic topic models \citep{blei2006dynamic},
latent feature models \citep{griffiths2007ibp},
hidden
Markov models for sequences \citep{rabiner1986introduction}
and trees \citep{crouse1998hmt},
linear dynamical system models \citep{shumway1982emForLDS,ghahramani1996estimationForLDS},
stochastic block models for 
relational data
\citep{wang1987stochasticBlockmodels,kemp2006irm},
 and many more.

\begin{figure}
\begin{tabular}{c c c}
\includegraphics[width=0.3\textwidth]{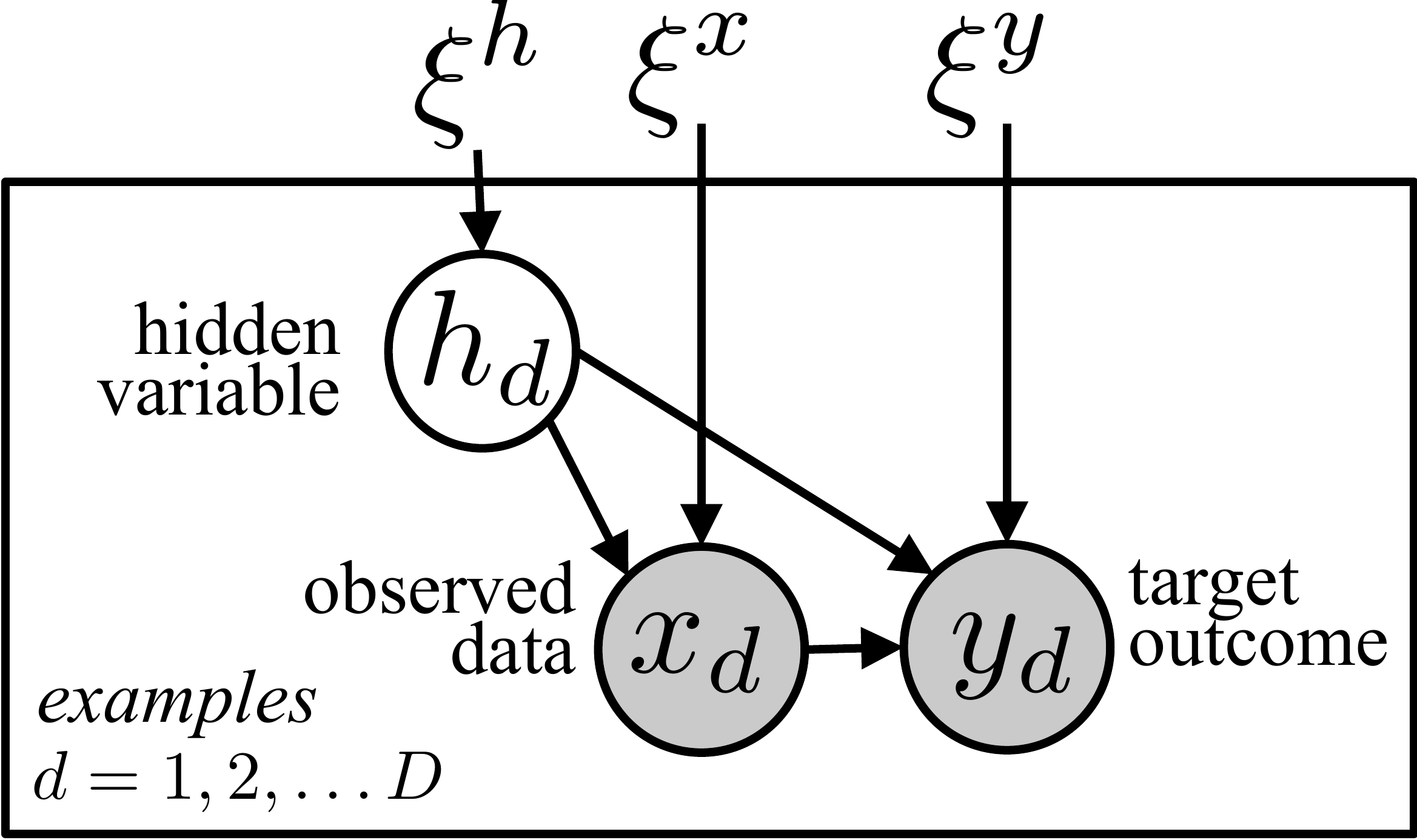}
&
\includegraphics[width=0.3\textwidth]{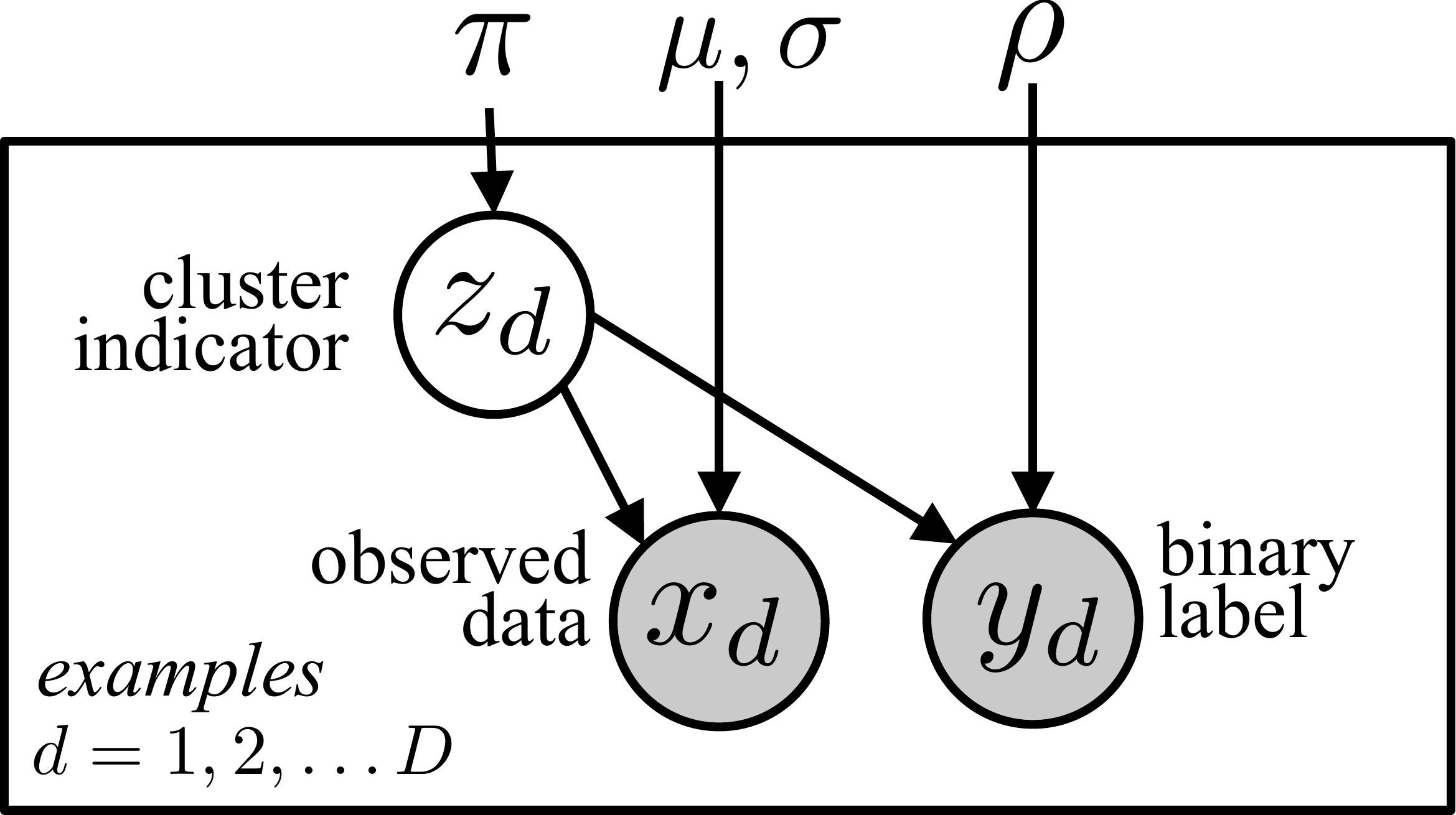}
&
\includegraphics[width=0.3\textwidth]{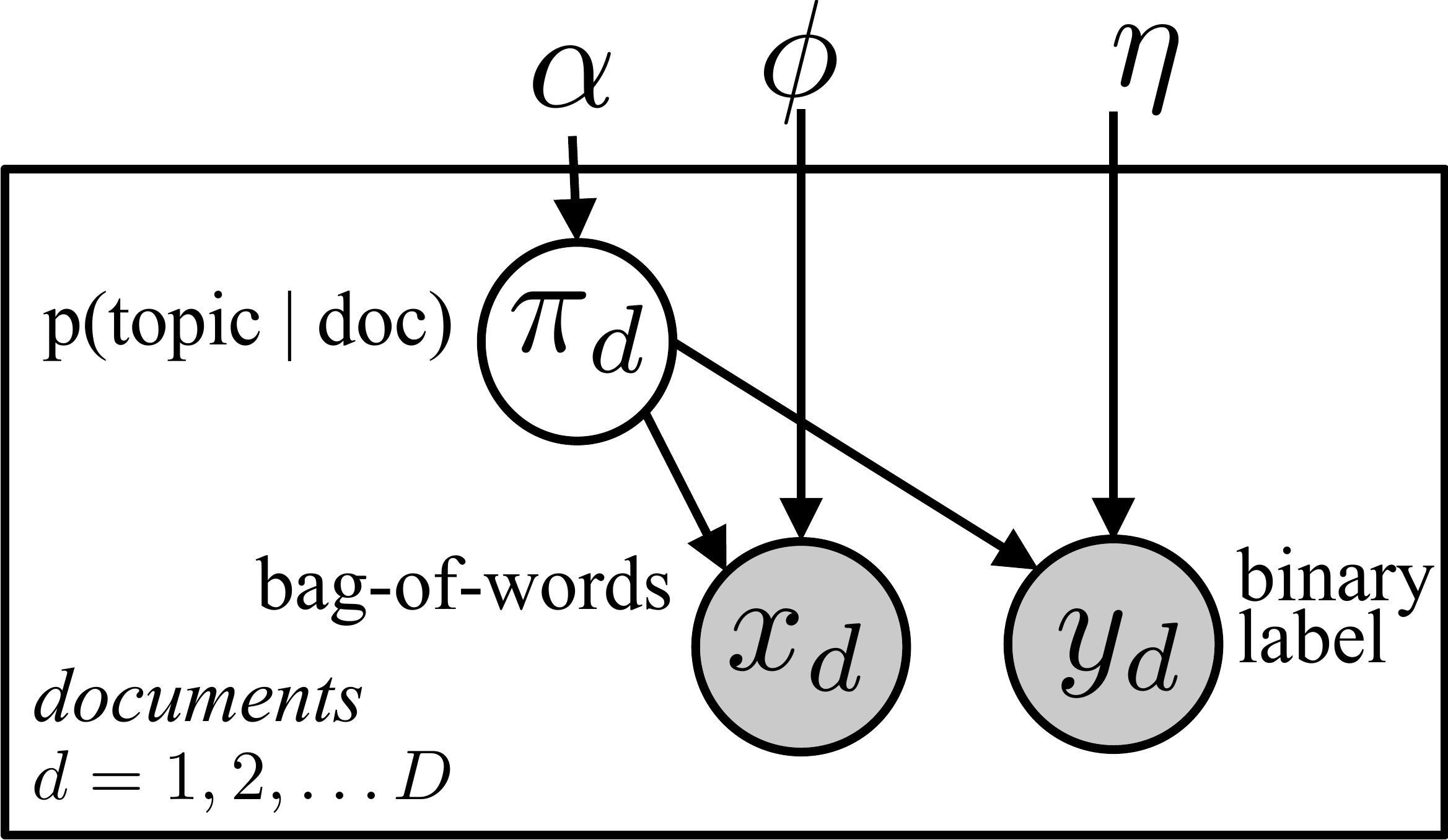}
\\
(a) General model
&
{\small
    (b) Supervised mixture (Sec.~\ref{sec:mixtures})
}
&
{\small
    (c) Supervised LDA (Sec.~\ref{sec:topics})
}
\end{tabular}
\caption{
Graphical models for downstream supervised LVMs 
amenable to prediction-constrained training.
}
\label{fig:model_diagrams}
\end{figure}

The broad family of latent variable model we consider is illustrated in Fig.~\ref{fig:model_diagrams}.
We assume an observed dataset of $D$ paired observations $\{x_d, y_d\}_{d=1}^D$. We refer
to $x_d$ as data and $y_d$ as labels or targets, with the understanding that in
intended applications, we can easily access some new data $x_d$ but
often need to predict $y_d$ from $x_d$. For example, the pairs $x_d,
y_d$ may be text documents and their accompanying class labels, or
images and accompanying scene categories, or patient medical histories
and their accompanying diagnoses. We will often refer to each observation (indexed by $d$) as a \emph{document}, since we are motivated in part by topic models, but we emphasize that our work is directly applicable to many other LVMs and data types.

We assume that each of the exchangeable data pairs $d$ is generated independently by the
model via its own \emph{hidden variable} $h_d$. 
For a simple mixture model, $h_d$ is an integer indicating the associated data cluster. For more complex members of our family like topic models, $h_d$ may be a set of several document-specific hidden variables. The generative process for the random variables $h_d, x_d, y_d$ local to document $d$ unfolds in three
steps: generate $h_d$ from some prior $P$, generate $x_d $ given $h_d$
according to some distribution $F$, and finally generate $y_d$ given both $x_d$ and $h_d$ from some distribution $G$. The joint density for document~$d$ then factorizes as 
\begin{align}
\label{eq:hxy_joint_density}
p(x_d, y_d, h_d \mid \xi) = 
    p(h_d \mid \xi^h)
    f(x_d \mid h_d, \xi^x)
    g(y_d \mid x_d, h_d, \xi^y).
\end{align}
We assume the generating distributions $P, F, G$ have parameterized probability density functions $p, f, g$ which can be easily evaluated and differentiated.
The global parameters $\xi^h$, $\xi^x$, and $\xi^y$ specify each
density.
When training our model, we treat the global parameters $\xi = [\xi^x, \xi^y, \xi^h]$ as random variables with associated prior density $p_0(\xi)$. 

Our chosen model family is an example of a \emph{downstream} LVM: the core assumption of Eq.~\eqref{eq:hxy_joint_density} is that the generative process produces both observed data $x_d$ and targets $y_d$ conditioned on the hidden variable $h_d$. 
In contast, \emph{upstream} models 
such as Dirichlet-multinomial regression~\citep{mimno2008dmr},
DiscLDA~\citep{lacoste2009disclda},
and labeled LDA~\citep{ramage2009labeled}
assume that observed labels $y_d$ are generated first, and then combined with hidden variables $h_d$ to produce data $x_d$.
For upstream models, inference is challenging when labels are missing. For example, in downstream models $p(h_d \mid x_d)$ may be computed by omitting factors containing $y_d$, while upstream models must explicitly integrate over all possible $y_d$.
Similarly, upstream prediction of labels $y_d$ from data $x_d$ is more complex than for downstream models.
That said, our predictively constrained framework could also be used to produce novel learning algorithms for upstream LVMs.  

Given this general model family, there are two core problems of
interest. The first is global parameter learning: estimating values
or approximate posteriors for $\xi$ given training data $\{x_d,
y_d\}$.  The second is local prediction: estimating the target $y_d$
given data $x_d$ and model parameters $\xi$.

\subsection{Regularized Maximum Likelihood Optimization for Training Global Parameters}
A classical approach to estimating $\xi$ would be to maximize the marginal likelihood of the training data $x$ and targets $y$, integrating over the hidden variables $h$.  This is equivalent to minimizing the following objective function:
\begin{align}
\label{eq:ml_objective}
\min_\xi
\quad &- \Bigg[
	\sum_{d=1}^D \log p(x_d, y_d \mid \xi^h, \xi^x, \xi^y)
    \Bigg] + R(\xi),
\\ \notag
p(x_d, y_d \mid \xi^h, \xi^x, \xi^y) 
    &= \int
            p(h_d \mid \xi^h)
            f(x_d \mid h_d, \xi^x)
            g(y_d \mid x_d, h_d, \xi^y)
            \;d h_d.
\end{align}
Here, $R(\xi)$ denotes a (possibly uninformative) regularizer for the global
parameters. If $R(\xi) = - \log p_0(\xi)$ for some prior density function $p_0(\xi)$, Eq.~\eqref{eq:ml_objective} is equivalent to \emph{maximum a posteriori} (MAP) estimation of $\xi$.

One problem with standard ML or MAP training is that the inputs $x_d$ and 
targets $y_d$ are modeled in a perfectly symmetric fashion.  We could equivalently concatenate $x_d$ and $y_d$ to form one larger variable, and use standard unsupervised learning methods to find a joint representation.  However, because practical models are typically misspecified and only approximate the generative process of real-world data, solving this objective can lead to solutions that are not matched to the
practitioner's goals.  We care much more about predicting patient
mortality rates than we do about estimating past incidences of routine
checkups. Especially because inputs $x_d$ are usually higher-dimensional than targets $y_d$, conventionally trained LVMs may have poor predictive performance.

\subsection{Prediction-Constrained Optimization for Training Global Parameters}
As an alternative to maximizing the joint likelihood, we consider a \emph{prediction-constrained} objective, where we wish to find
the best possible generative model for data $x$ that meets some
quality threshold for prediction of targets $y$ given $x$. A natural quality
threshold for our probabilistic model is to require that the sum of
log conditional probabilities $p(y_d \mid x_d, \xi)$ must exceed 
some scalar value $L$.  This leads to the following constrained optimization problem:
\begin{align}
\label{eq:pc_objective_constrained_format}
\min_{\xi} 
\quad &- \Bigg[ \sum_{d=1}^D
        \log p(x_d \mid \xi^x, \xi^h) 
    \Bigg]
    + R(\xi),
\\ \notag
\mbox{subject to~} \quad 
    &- \sum_{d=1}^D
    \log p(y_d \mid x_d, \xi^h, \xi^x, \xi^y) \leq L.
\end{align}
We emphasize that the conditional probability $p(y_d \mid x_d, \xi)$ \emph{marginalizes} the hidden variable $h_d$:
\begin{align}
p(y_d \mid x_d, \xi^h, \xi^x, \xi^y)
    &= \int g(y_d \mid x_d, h_d, \xi^y) p(h_d \mid x_d, \xi^h, \xi^x) \;dh_d.
\end{align}
This marginalization allows us to make predictions for $y_d$ that 
correctly account for our uncertainty in $h_d$ given $x_d$, and
importantly, given \emph{only} $x_d$. 
If our goal is to predict $y_d$ given $x_d$, then we cannot train our model assuming $h_d$ is informed by both $x_d$ and $y_d$.  

Lagrange multiplier theory tells us that any solution of the constrained problem in Eq.~\eqref{eq:pc_objective_constrained_format} as also a solution to the unconstrained optimization problem
\begin{align}
\label{eq:pc_objective_unconstrained_format}
\min_{\xi} \quad &
    - \Bigg[ \sum_{d=1}^D \log p(x_d \mid \xi^h, \xi^x)
      \Bigg]
    - \lambda \Bigg[
        \sum_{d=1}^D 
              \log p(y_d \mid x_d, \xi^h, \xi^x, \xi^y)
        \Bigg]
    + R(\xi),
\end{align}
for some scalar Lagrange multiplier $\lambda > 0$.  For each distinct value of $\lambda$, the solution to Eq.~\eqref{eq:pc_objective_unconstrained_format} also solves the constrained problem in Eq.~\eqref{eq:pc_objective_constrained_format} for a particular threshold $L$. While the mapping between $\lambda$ and $L$ is monotonic, it is not constructive and lacks a simple parametric form.

We define the optimization problem in
Eq.~\eqref{eq:pc_objective_unconstrained_format} to be our
\emph{prediction-constrained} (PC) training objective. This objective
directly encodes the asymmetric relationship between data
$x_d$ and labels $y_d$ by prioritizing prediction of $y_d$ from $x_d$
when $\lambda > 1$.  
This contrasts with the \emph{joint maximum likelihood} objective in Eq.~\eqref{eq:ml_objective} which treats these variables symmetrically, and (especially when $x_d$ is high-dimensional) may not accurately model the predictive density $p(y_d \mid x_d)$.
In the special case where $\lambda = 1$, the PC objective of
Eq.~\eqref{eq:pc_objective_unconstrained_format} reduces to the ML
objective of Eq.~\eqref{eq:ml_objective}.

\subsubsection{Extension: Constraints on a general expected loss}
Penalizing aggregate log predictive probability is sensible for many problems, but for some applications other loss functions are more appropriate.  More generally, we can penalize the \emph{expected loss} between the true labels $y_d$ and predicted labels $\hat{y}(x_d, h_d, \xi^y)$ under the LVM posterior $p(h_d \mid x_d, \xi)$:
\begin{align}
\label{eq:pc_objective_loss_format}
\min_{\xi} 
\quad &- \Bigg[ \sum_{d=1}^D
        \log p(x_d \mid \xi^x, \xi^h) 
    \Bigg]
    + R(\xi),
\\ \notag
\mbox{subject to~} \quad 
    &\sum_{d=1}^D \mathbb{E}_{h_d}
        [
        \lossyxh \mid x_d,\xi
        ]
    \leq L.
\end{align}
This more general approach allows us to incorporate classic non-probabilistic loss functions like the hinge loss or epsilon-insensitive loss, or to penalize errors asymmetrically in classification problems, when measuring the quality of predictions.
However, for this paper, our algorithms and experiments focus on the probabilistic loss formulation in Eq.~\eqref{eq:pc_objective_unconstrained_format}.

\subsubsection{Extension: Prediction constraints for individual data items}
In Eq.~\eqref{eq:pc_objective_constrained_format}, we defined our
prediction quality constraint using the sum (or equivalently, the average) of the
document-specific losses $\log p(y_d \mid x_d, \xi^h, \xi^x, \xi^y)$.  An alternative, more stringent training object would enforce separate prediction constraints for each document:
\begin{align}
\label{eq:pc_objective_individual_loss_format}
\min_{\xi} 
\quad &- \Bigg[ \sum_{d=1}^D
        \log p(x_d \mid \xi^x, \xi^h) 
    \Bigg]
    + R(\xi),
\\ \notag
\mbox{subject to~} \quad 
    &- \log p(y_d \mid x_d, \xi^h, \xi^x, \xi^y) \leq L_d \quad \text{for all } d. 
\end{align}
This modified optimization problem would generalize Eq.~\eqref{eq:pc_objective_unconstrained_format} by allocating a distinct Lagrange multiplier weight 
$\lambda_d$ for each observation $d$.  Tuning these weights would require more sophisticated optimization algorithms, a topic we leave for future research.

\subsubsection{Extension: Semi-supervised prediction constraints for data with missing labels}
In many applications, we have a dataset of $D$ observations $\{x_d\}_{d=1}^D$
for which only a subset $\mathcal{D}^y \subset \{1, 2, \ldots D\}$ have observed labels $y_d$; the remaining labels are unobserved.
For semi-supervised learning problems like this, we generalize Eq.~\eqref{eq:pc_objective_constrained_format} to only enforce the label prediction constraint for the documents in $\mathcal{D}^y$, so that the PC objective of Eq.~\eqref{eq:pc_objective_constrained_format} becomes:
\begin{align}
\label{eq:pc_objective_semisupervised_constrained} 
\min_{\xi} 
\quad &- \Bigg[ \sum_{d=1}^D
        \log p(x_d \mid \xi^x, \xi^h) 
    \Bigg]
    + R(\xi),
\\ \notag
\mbox{subject to~} \quad 
    &- \sum_{d : \mathcal{D}^y}
    \log p(y_d \mid x_d, \xi^h, \xi^x, \xi^y) \leq L.
\end{align}
In general, the value of $L$ will need to be adapted based on the amount of labeled data.  In the unconstrained form
\begin{align}
\label{eq:pc_objective_semisupervised_unconstrained}
\min_{\xi} \quad &
    - \Bigg[ \sum_{d=1}^D \log p(x_d \mid \xi^h, \xi^x)
      \Bigg]
    - \lambda \Bigg[
      \sum_{d : \mathcal{D}^y}
              \log p(y_d \mid x_d, \xi^h, \xi^x, \xi^y)
        \Bigg]
    + R(\xi),
\end{align}
as the fraction of labeled data $b = \frac{|\mathcal{D}^y|}{D}$ gets smaller, we will need a much larger Lagrange multiplier $\lambda$ to uphold the \emph{same} average quality in predictive performance.
This occurs simply because as $b$ gets smaller, the data likelihood term $\log p(x)$ will 
continue to get larger in relative magnitude compared to the label prediction term $\log p(y \mid x)$. 

\subsection{Relationship to Other Supervised Learning Frameworks}
\label{sec:problems}

While the definition of the PC training objective in 
Eq.~\eqref{eq:pc_objective_unconstrained_format} is straightforward, 
it has desirable features that are not shared by other supervised training objectives for downstream LVMs.
In this section we contrast the PC objective with several other approaches, often comparing to methods from the topic modeling literature to give concrete alternatives.

\subsubsection{Advantages over standard joint likelihood training}
For our chosen family of supervised downstream LVMs,
the most standard training method is to find a point estimate of global
parameters $\xi$ that maximizes the (regularized) joint log-likelihood $\log
p(x, y \mid \xi)$ as in Eq.~\eqref{eq:ml_objective}.  Related Bayesian
methods that approximate the posterior distribution $p(\xi \mid x, y)$,
such as variational methods~\citep{wainwright2008variational} and
Markov chain Monte Carlo methods~\citep{andrieu2003introMCMC}, estimate moments of
the same \emph{joint} likelihood (see Eq.~\eqref{eq:hxy_joint_density}) relating hidden variables $h_d$ to data $x_d$ and labels $y_d$.

For example, supervised LDA \citep{blei2007sLDA,wang2009simultaneous} 
learns latent topic assignments $h_d$ by optimizing the joint probability of 
bag-of-words document representations $x_d$ and document labels $y_d$.
One of several problems with this joint likelihood objective is \emph{cardinality mismatch}: the relative sizes of the random
variables $x_d$ and $y_d$ can reduce predictive performance. 
In particular, if $y_d$ is a one-dimensional binary label but $x_d$ is a high-dimensional word count vector,
the optimal solution to Eq.~\eqref{eq:ml_objective} will often
be indistinguishable from the solution to the \emph{unsupervised}
problem of modeling the data $x$ alone.  Low-dimensional labels can
have neglible impact on the joint density compared to the
high-dimensional words $x_d$, causing learning to ignore subtle features that are critical for the prediction of $y_d$ from $x_d$.  
Despite this issue, recent work continues to use this training objective
\citep{wang2014spectral,ren2017spectral}.

\subsubsection{Advantages over maximum conditional likelihood training}
Motivated by similar concerns about joint likelihood training, \citet{jebara1999cem} introduce a method to explicitly optimize the conditional likelihood 
$\log p(y \mid x, \xi)$ for a particular LVM, the Gaussian mixture model.  They replace the conditional likelihood with a more tractable lower bound, and then monotonically increase this bound via a coordinate ascent algorithm they call \emph{conditional expectation maximization} (CEM).
\citet{chen2015bplda} instead use a variant of backpropagation to optimize the conditional likelihood of a supervised topic model.

One concern about the conditional likelihood objective is that it \emph{exclusively} focuses on the prediction task; it need not lead to good models of the data $x$, and cannot incorporate unlabeled data.  In contrast, our prediction-constrained approach allows a principled tradeoff between optimizing the marginal likelihood of data and the conditional likelihood of labels given data.

\subsubsection{Advantages over label replication}
We are not the first to notice that high-dimensional data $x_d$ can
swamp the influence of low-dimensional labels $y_d$.  Among
practitioners, one common workaround to this imbalance is to retain the
symmetric maximum likelihood objective of Eq.~\eqref{eq:ml_objective},
but to \emph{replicate} each label $y_d$ as if it were observed $r$ times per document:
$\{y_d, y_d, \ldots, y_d \}$.  Applied to supervised LDA, label replication
leads to an alternative \emph{power sLDA} topic model~\citep{zhang2014howToSuperviseTopicModels}.

Label replication still leads to nearly the same per-document joint
density as in Eq.~\eqref{eq:hxy_joint_density}, except that the likelihood density
is raised to the $r$-th power: $g(y_d \mid x_h, h_d, \xi^y)^r$. 
While label replication can better ``balance'' 
the relative sizes of $x_d$ and $y_d$ when $r \gg 1$, performance gains over standard supervised LDA are often negligible~\citep{zhang2014howToSuperviseTopicModels}, because this approach does not address the \emph{assymmetry issue}.  To see why,
we examine the label-replicated training objective:
\begin{align}
\label{eq:ml_objective_replicated}
\min_\xi
\quad &- 
	\sum_{d=1}^D 
	\log 
\Bigg[
    \int
            p(h_d \mid \xi^h)
            f(x_d \mid h_d, \xi^x)
            g(y_d \mid x_d, h_d, \xi^y)^r
            \;d h_d
\Bigg]
    + R(\xi).
\end{align}
This objective does not contain any direct penalty on the predictive
density $p(y_d \mid x_d)$, which is the fundamental idea of our
prediction-constrained approach and a core term in the objective of
Eq.~\eqref{eq:pc_objective_unconstrained_format}. Instead, only the
symmetric joint density $p(x, y)$ is maximized, with training assuming
both data $x$ and replicated labels $y$ are present.  
It is easy to find examples where the
optimal solution to this objective performs poorly on the
target task of predicting $y$ given only $x$, because the training has
not directly prioritized this asymmetric prediction.  In later
sections such as the case study in
Fig.~\ref{fig:mix_toy_1d_asym_example}, we provide intuition-building
examples where maximum likelihood joint training with label
replication fails to give good prediction performance for \emph{any}
value of the replication weight, while our PC approach can do better
when $\lambda$ is sufficiently large.

\paragraph{Example: Label replication may lead to poor predictions.}
Even when the number of replicated labels $r \to \infty$, the optimal solution to the label-replicated training objective of Eq.~\eqref{eq:ml_objective_replicated} may be suboptimal for the prediction of $y_d$ given $x_d$.  To demonstrate this, we consider a toy example involving two-component Gaussian mixture models.

Consider a one-dimensional data set consisting of six evenly spaced points,
$x = \{1,2,3,4,5,6\}$.  The three points where $x \in \{2,4,5\}$ have positive labels $y = 1$, while the rest have negative labels $y=0$.  Suppose our goal is to fit a
mixture model with two Gaussian components to these data, assuming minimal
regularization (that is, sufficient only to prevent the probabilities of clusters and
targets from being exactly 0 or 1).
Let $h_d \in \{0,1\}$ indicate the (hidden) mixture component for $x_d$.

If $r \gg 1$, the $g(y_d \mid x_d,h_d,\xi^y)^r$ term will dominate in
Eq.~\eqref{eq:ml_objective_replicated}.  This term can be optimized by
setting $h_d = y_d$, and the probability of $y_d = 1$ to close to 0 or
1 depending on the cluster.  In particular, we choose $p(y_d=1 \mid h_d=0) = 0.0001$ and
$p(y_d=1 \mid h_d=1) = 0.9999$.  If one computes the maximum likelihood
solution to the remaining parameters given these assignments of $h_d$,
the resulting labels-from-data likelihood equals $\sum_{d=1}^D \log p(y_d \mid x_d) = -3.51$, and two points are misclassified.  Misclassification occurs because the two clusters have significant overlap.

However, there exists an alternative two-component mixture model that yields better labels-given-data likelihood and makes fewer mistakes. We set the cluster centers to
$\mu_0 = 2.0$ and $\mu_1 = 4.5$, and the cluster variances to $\sigma_0 = 5.0$ and $\sigma_1 = 0.25$.
Under this model, we get a labels-given-data likelihood of
$\sum_{d=1}^D \log p(y_d \mid x_d) = -2.66$, and only one point is
misclassified.  This solution achieves a lower misclassification rate
by choosing one narrow Gaussian cluster to model the adjacent positive points $x \in
\{4,5\}$ correctly, while making no attempt to capture the positive point at $x = 2$.
Therefore, the solution to
Eq.~\eqref{eq:ml_objective_replicated} is suboptimal for making predictions about $y_d$ given $x_d$.

This counter-example also illustrates the intuition behind why the
replicated objective fails: increasing the replicates of $y_d$ forces
$h_d$ to take on a value that is predictive of $y_d$ during training,
that is, to get $p(y_d \mid h_d)$ as close to 1 as possible.  However, there are no guarantees on $p(h_d \mid x_d)$ which is necessary for predicting $y_d$ given $x_d$.  See Fig.~\ref{fig:mix_toy_1d_asym_example} for an additional in-depth example.

\subsubsection{Advantages over posterior regularization}
The \emph{posterior regularization} (PR) framework introduced by \citet{gracca2008posteriorconstraints}, and later refined in \citet{ganchev2010posteriorconstraints}, is notable early work which applied explicit performance constraints to latent variable model objective functions.
Most of this work focused on models for only two local random variables: data $x_d$ and hidden variables $h_d$, without any explicit labels $y_d$.
Mindful of this, we can naturally express the PR objective in our notation, explaining data $x$ explicitly via an objective function and incorporating labels $y$ only later in the performance constraints.

The PR approach begins with the same overall goals of the expectation-maximization treatment of maximum likelihood inference: frame the problem as estimating an approximate posterior $q(h_d \mid \hat{v}_d)$ for each latent variable set $h_d$, such that this approximation is as close as possible in KL divergence to the real (perhaps intractable) posterior $p(h_d \mid x_d, y_d, \xi)$. Generally, we select the density $q$ to be from a tractable parametric family with free parameters $\hat{v}_d$ restricted to some parameter space $\hat{v}_d \in \mathcal{V}$ which makes $q$ a valid density. This leads to the objective
\begin{align}
\label{eq:ml_objective_lower_bound}
    \min_{\xi, \{\hat{v}_d \}_{d=1}^D}
    \quad
    & R(\xi) - \sum_{d=1}^D \mathcal{L}(x_d, \hat{v}_d, \xi),
\\
\mathcal{L}(x_d, \hat{v}_d, \xi) &\triangleq 
    \mathbb{E}_q \Big[
        \log p(x_d, h_d \mid \xi)
        - \log q(h_d \mid \hat{v}_d)
        \Big]
        \leq 
    \log p(x_d | \xi).
\end{align}
Here, the function $\mathcal{L}$ is a strict \emph{lower bound} on the data likelihood $\log p(x_d \mid \xi)$ of Eq.~\eqref{eq:ml_objective}.
The popular EM algorithm optimizes this objective via coordinate descent steps that alternately update variational parameters $\hat{v}_d$ and model parameters $\xi$.
The PR framework of \citet{gracca2008posteriorconstraints} adds additional constraints to the approximate posterior $q(h_d \mid \hat{v}_d)$ so that some additional loss function of interest, over both observed and latent variables, has bounded value under the distribution $q(h_d)$:
\begin{align}
\label{eq:PR_constraint}
\mbox{Posterior Regularization (PR):}
\quad
    \mathbb{E}_{q(h_d)}
    \Big[ \lossyxh
    \Big] \leq L.
\end{align}
For our purposes, one possible loss function could be the negative log likelihood for the label $y$: $\lossyxh = - \log g(y_d \mid x_d, h_d, \xi^y)$. It is informative to directly compare the PR constraint above with the PC objective of Eq.~\eqref{eq:pc_objective_loss_format}. Our approach directly constrains the expected loss under the \emph{true} hidden-variable-from-data posterior $p(h_d | x_d)$: 
\begin{align}
\label{eq:PC_constraint}
\mbox{Prediction Constrained (PC):}
\quad
    \mathbb{E}_{p(h_d|x_d)}
    \Big[ \lossyxh
    \Big] \leq L.
\end{align}
In contrast, the PR approach in Eq.~\eqref{eq:PR_constraint}
constrains the expectation under the \emph{approximate posterior}
$q(h_d)$.  This posterior does not have to stay close to \emph{true}
hidden-variable-from-data posterior $p(h_d | x_d)$. Indeed, when we
write the PR objective in unconstrained form with Lagrange multiplier
$\lambda$, and assume the loss is the negative label log-likelihood, we have:
\begin{align}
    \min_{\xi, \{\hat{v}_d \}_{d=1}^D}
    - & \mathbb{E}_q \Bigg[
        \sum_{d=1}^D \log p(x_d, h_d \mid \xi)
        + \lambda \log g(y_d \mid x_d, h_d, \xi^y)
        - \log q(h_d | \hat{v}_d)
        \Bigg] + R(\xi)
\end{align}
Shown this way, we reach a surprising conclusion: the PR objective reduces to a lower bound on the symmetric joint likelihood with labels replicated $\lambda$ times.  Thus, it will inherit all the problems of label replication discussed above, as the optimal training update for $q(h_d)$ incorporates information from \emph{both} data $x_d$ and labels $y_d$. However, this does \emph{not} train the model to find a good approximation of $p(h_d \mid x_d)$, which we will show is critical for good predictive performance.

\subsubsection{Advantages over maximum entropy discrimination and regularized Bayes}
Another key thread of related work putting constraints on approximate
posteriors is known as \emph{maximum entropy discrimination} (MED),
first published in \citet{jaakkola1999med} with further details in followup work~\citep{jaakkola1999MEDtechreport,jebara2001medthesis}.
This approach was developed for training discriminative models
without hidden variables, where the primary innovation was showing how
to manage uncertainty about parameter estimation under max-margin-like
objectives.
In the context of LVMs, this MED work differs from standard EM optimization 
in two important and separable ways. First, it estimates a posterior for
global parameters $q(\xi)$ instead of a simple point estimate. Second, it enforces a margin constraint on label prediction, rather than just maximizing log probability of labels.
We note briefly that \citet{jaakkola1999MEDtechreport} did consider a MED objective for
\emph{unsupervised} latent variable models (see their Eq.~48), where the constraint is directly on the expectation of the lower-bound of the log data likelihood.
The choice to constrain the data likelihood is fundamentally different from constraining the labels-given-data loss, which was not done for LVMs by the original MED work yet is more aligned with our focus with high-quality predictions.


The key application MED to supervised LVMs has been \cite{zhu2012medlda}'s MED-LDA, an extension of the LDA topic model based on a MED-inspired training objective.
Later work developed similar objectives for other LVMs under the broad name of \emph{regularized Bayesian inference}~\citep{zhu2014regbayes}.
To understand these objectives, we focus on \citet{zhu2012medlda}'s original unconstrained training objectives for MED-LDA for both regression
(Problem 2, Eq.~8 on p.~2246) and classification (Problem 3, Eq.~19 on
p.~2252), which can be fit into our notation\footnote{
We note an irregularity between the classification and regression formulation of MED-LDA published by ~\citet{zhu2012medlda}: 
while classification-MED-LDA included labels $y$ only the loss term, the regression-MED-LDA included \emph{two} terms in the objective that penalize reconstruction of $y$: 
one inside the likelihood bound term $\mathcal{L}$ using a Gaussian likelihood $G$ as well as inside a separate epsilon-insensitive loss term. Here, we assume that only the loss term is used for simplicity.}
as follows:
\begin{align*}
\min_{
    q(\xi),
    \{\hat{v}_d \}_{d=1}^D}
\quad
    \mbox{KL}( q(\xi) || p_0(\xi) )
    - \mathbb{E}_{q(\xi)}
        \Big[ \sum_{d=1}^D \mathcal{L}(x_d, \hat{v}_d, \xi) \Big]
    + C \sum_{d=1}^D
        \text{loss}(y_d, \mathbb{E}_{q(\xi, h_d)}
        [ \hat{y}_d( x_d, h_d, \xi) ] )
\end{align*}
Here $C > 0$ is a scalar emphasizing how important the loss function
is relative to the unsupervised problem,
$p_0(\xi)$ is some prior distribution on global parameters, and
$\mathcal{L}(x_d, \hat{v}_d, \xi)$ is the same lower bound as in Eq.~\eqref{eq:ml_objective_lower_bound}. We can make this objective more comparable to our earlier objectives by performing point estimation of $\xi$ instead of posterior approximation, which
is reasonable in moderate to large data regimes, as the posterior for the global parameters $\xi$ will concentrate.
This choice allows us to focus on our core question of how to define an objective that balances data $x$ and labels $y$, rather than the separate question of managing uncertainty during this training.
Making this simplification by substituting point estimates for expectations, with the KL divergence regularization term reducing to $R(\xi) = - \log p_0( \xi )$, and the MED-LDA objective becomes:
\begin{align}
\label{eq:med_loss_of_E_objective}
\min_{
    \xi,
    \{\hat{v}_d \}_{d=1}^D}
\quad
    R(\xi)
    - \sum_{d=1}^D \mathcal{L}(x_d, \hat{v}_d, \xi)
    + C \sum_{d=1}^D
        \text{loss}(
            y_d,
        \mathbb{E}_{q(h_d)} [
            \hat{y}_d( x_d, h_d, \xi)
        ]
            ).
\end{align}
Both this objective and \citet{gracca2008posteriorconstraints}'s PR framework consider expectations over the approximate posterior $q(h_d)$, rather than our choice of the data-only posterior $p(h_d | x_d)$. However, the key difference between MED-LDA and the PR objectives is that the MED-LDA objective computes the loss of an expected prediction ($\text{loss}(y_d, \mathbb{E}_q[ \hat{y}_d ])$), while the earlier PR objective in Eq.~\eqref{eq:PR_constraint} penalizes the full expectation of the loss ($\mathbb{E}_{q(h_d)}[ \text{loss}(y_d,\hat{y}_d )]$). Earlier MED work~\citep{jaakkola1999MEDtechreport} also suggests using an expectation of the loss, $\mathbb{E}_{q(\xi, h_d)}[ \text{loss}(y_d,\hat{y}_d(x_d, h_d, \xi))]$.
Decision theory argues that the latter choice is preferable when possible, since it should lead to decisions that better minimize loss under uncertainty. We suspect that MED-LDA chooses the former only because it leads to more tractable algorithms for their chosen loss functions. 

Motivated by this decision-theoretic view, we consider modifying the MED-LDA objective of Eq.~\eqref{eq:med_loss_of_E_objective} so that we take the full expectation of the loss. 
This swap can also be justified by assuming the loss function is \emph{convex}, as are both the epsilon-insensitive loss and the hinge loss used by MED-LDA, so that Jensen's inequality may be used to bound the objective in Eq.~\eqref{eq:med_loss_of_E_objective} from above.
The resulting training objective is:
\begin{align}
\label{eq:med_E_of_loss_objective}
\min_{
    \xi,
    \{\hat{v}_d \}_{d=1}^D}
\quad
    R(\xi)
    - \sum_{d=1}^D \mathcal{L}(x_d, \hat{v}_d, \xi)
    + C \sum_{d=1}^D
        \mathbb{E}_{q(h_d)} \Big[
        \text{loss}(
            y_d,
            \hat{y}_d( x_d, h_d, \xi)
            )
        \Big].
\end{align}
In this form, we see that we have recovered the symmetric maximum likelihood objective with label replication from Eq.~\eqref{eq:ml_objective_replicated}, 
with $y$ replicated $C$ times. Thus, even this MED effort fails to properly handle the asymmetry issue we have raised, possibly leading to poor generalization performance.


\subsection{Relationship to Semi-supervised Learning Frameworks}

Often, semi-supervised training is performed via optimization of the joint likelihood $\log p(x, y \mid \xi)$, using the EM algorithm to impute missing data~\citep{nigam1998semisuperEM}. 
Other work falls under the thread of ``self-training'', where a model trained on labeled data only is used to label additional data and then retrained accordingly.  \citet{chang2007semisupervizedwithconstraint} incorporated constraints into semi-supervised self-training of an upstream hidden Markov model (HMM). 
Starting with just a small labeled dataset, they iterate between two steps:
(1) train model parameters $\xi$ via maximum likelihood estimation on the fully labeled set,
and (2) expand and revise the fully labeled set via a constraint-driven approach.
Given several candidate labelings $y_d$ for some example, their step 2 reranks these to prefer those that obey some soft constraints (for example, in a bibliographic labeling task, they require the ``title'' field to always appear once).
Importantly, however, this work's subprocedure for training from fully labeled data is a symmetric maximum likelihood objective, while our PC approach more directly encodes the asymmetric structure of prediction tasks.

Other work deliberately specifies prior domain knowledge about label distributions, and penalizes models that deviate from this prior when predicting on unlabeled data.
\citet{mann2010generalized} propose \emph{generalized expectation} (GE) constraints, which extend their earlier \emph{expectation regularization} (XR) approach~\citep{mann2007expectationRegularization}.
This objective has two terms:
a conditional likelihood objective, and a new regularization term comparing model predictions to some weak domain knowledge:
\begin{align}
\log p(y | x, \xi) - \lambda \Delta( \hat{Y}(x, \xi), Y_H ).
\end{align}
Here, $Y_H$ indicates some expected domain knowledge about the overall labels-given-data distribution, while $\hat{Y}(x, \xi)$ is the predicted labels-given-data distribution 
under the current model. The distance function $\Delta$, weighted by $\lambda > 0$, penalizes predictions that deviate from the domain knowledge. Unlike our PC approach, this objective focuses exclusively on the label prediction task and does not at all incorporate the notion of generative modeling.

\section{Case Study: Prediction-constrained Mixture Models}
\label{sec:mixtures}
We now present a simple case study
applying prediction-constrained training to supervised mixture models.
Our goal is to illustrate the benefits of
our prediction-constrained approach in a situation where the
marginalization over $h_d$ in
Eq.~\eqref{eq:pc_objective_unconstrained_format} can be computed exactly in closed form. This allows direct comparison of our proposed PC training objective to alternatives like maximum likelihood, without worry about how approximations needed to make inference tractable affect either objective.

Consider a simple supervised mixture model which generates data pairs $x_d, y_d$, as illustrated in Fig.~\ref{fig:model_diagrams}(b). 
This mixture model assumes there are $K$ possible discrete hidden states, and that the only hidden variable at each data point $d$ is an indicator variable: $h_d = \{ z_d \}$, 
where $z_d \in \{1, 2, \ldots K\}$
indicates which of the $K$ clusters point $d$ is assigned to.
For the mixture model, we parameterize the densities in Eq.~\eqref{eq:hxy_joint_density} as follows:
\begin{align}
\log p(z_d=k \mid \xi^h) &= \log \pi_k,
\\
\log p(x_d \mid z_d=k, \xi^x) &= 
    \log f( x_d \mid \xi^x_k ),
\\
\log p(y_d \mid x_d, z_d=k, \xi^y) &=
    \log g( y_d \mid x_d, \xi^y_{k} ).
\end{align}
The parameter set of the latent variable prior $P$ is simple: $\xi^h = \{\pi \}$, where $\pi$ is a vector of $K$ positive numbers that sum to one, representing the prior probability of each cluster.

We emphasize that the data likelihood $f$ and label likelihood $g$ are left in generic form since these are relatively modular: one could apply the mixture model objectives below with many different data and label distributions, so long as they have valid densities that are easy to evaluate and optimize for parameters $\xi^x, \xi^y$. Fig.~\ref{fig:model_diagrams}(b) happens to show the particular likelihood choices we used in our toy data experiments (Gaussian distribution for $F$, bernoulli distribution for $G$), but we will develop our PC training for the general case.
The only assumption we make is that each of the $K$ clusters has a separate parameter set:
$\xi^x = \{ \xi^x_k \}_{k=1}^K$ and
$\xi^y = \{ \xi^y_k \}_{k=1}^K$.

\paragraph{Related work on supervised mixtures.}
While to our knowledge, our prediction-constrained optimization objective is novel, there is a large related literature on applying mixtures to supervised problems where the practioner observes pairs of data covariates $x$ and targets $y$.
One line of work uses generative models with factorization structure like Fig.~\ref{fig:model_diagrams}, where each cluster $k$ has parameters for generating data $\xi^x_k$ and targets $\xi^y_k$. 
For example, \citet[Sec.~4.2]{ghahramani1993supervisedEMforMixtures} consider nearly the same model as in our toy experiments (except for using a categorical over labels $y$ instead of a Bernoulli). They derive an Expectation Maximization (EM) algorithm to maximize a lower bound on the symmetric joint log likelihood $\log p(x, y \mid \xi)$. Later applied work has sometimes called such models Bayesian profile regression when the targets $y$ are real-valued~\citep{molitor2010bayesianProfileRegression}. These efforts have seen broad extensions to generalized linear models especially in the context of Bayesian nonparametric priors like the Dirichlet process fit with MCMC sampling procedures~\citep{shahbaba2009nonlinearDPmix,hannah2011DPmixGLM,liverani2015premium}. However, none of these efforts correct for the assymmetry issues we have raised, instead simply using the symmetric joint likelihood.

Other work takes a more discriminative view of the clustering task.
\citet{krause2010discriminative}
develop an objective called Regularized Information maximization
which learns a conditional distribution for $y$ that preserves information from the data $x$.
Other efforts do not estimate probability densities at all, such as ``supervised clustering''~\citep{eick2004supervised}.
Many applications of this paradigm exist~\citep{
finley2005supervised,
al2006adapting,
dicicco2010machine,
peralta2013enhancing,
ramani2013improved,
grbovic2013supervised,
peralta2016proposal,
flammarion2016robust,
ismaili2016supervised,
yoon2016personalized,
dhurandhar2017uncovering}.

\subsection{Objective function evaluation and parameter estimation.}

\paragraph{Computing the data log likelihood.}
The marginal likelihood of a single data example $x_d$, marginalizing over the latent variable $z_d$, can be computed in closed form via the function:
\begin{align}
M^{x}(x_d, \pi, \xi^x) &\triangleq \log p(x_d \mid \pi, \xi^x) 
\\ \notag
&=  \log 
    \sum_{k=1}^K 
    \exp \!\Big(
        \log f(x_d \mid \xi^x_k) + \log \pi_k \Big).
\end{align}

\paragraph{Computing the label given data log likelihood.}
Similarly, the likelihood $p(y_d \mid x_d)$ of labels given data, marginalizing away the latent variable $z_d$, can be computed in closed form:
\begin{multline}
    M^{y|x}(y_d, x_d, \pi, \xi^x, \xi^y) \triangleq
    \log p(y_d \mid x_d, \pi, \xi^x, \xi^y)
    \\ \notag
    =
    \log \Bigg[
    \sum_{k=1}^K 
    \exp \!\Big(
        \log g(y_d \mid x_d, \xi^y_k) + 
        \log f(x_d \mid \xi^x_k) +
        \log \pi_k
        \Big) \Bigg]
        - M^x(x_d, \pi, \xi^x).
\end{multline}

\paragraph{PC parameter estimation via gradient descent.}
Our original unconstrained PC optimization problem in Eq.~\eqref{eq:pc_objective_unconstrained_format} can thus be formulated for mixture models using this closed form marginal probability functions $M$ and appropriate regularization terms $R$:
\begin{align}
\min_{
    \pi \in \Delta^K,
    ~\xi^x,
    ~\xi^y}
\quad 
- M^x(x_d, \pi, \xi^x) - \lambda M^{y|x}(y_d, x_d, \pi, \xi^x, \xi^y) + R(\xi).
\end{align}
We can practically solve this optimization objective via gradient descent. However, some parameters such as $\pi$ live in constrained spaces like the $K-$dimensional simplex. To handle this, we apply invertible, one-to-one transformations from these constrained spaces to unconstrained real spaces and apply standard gradient methods easily.

In practice, for training supervised mixtures we use the Adam gradient descent procedure \citep{kingma2014adam}, which requires specifying some baseline learning rate (we search over a small grid of 0.1, 0.01, 0.001) which is then adaptively scaled at each parameter dimension to improve convergence rates. We initialize parameters via random draws from reasonable ranges and run several thousand gradient update steps to achieve convergence to local optima. To be sure we find the best possible solution, we use many (at least 5, preferably more) random restarts for each possible learning rate and choose the one snapshot with the lowest training objective score.

\subsection{Toy Example: Why Asymmetry Matters}
\label{sec:mix_toy_1d_asym}
\begin{figure}[!t]
\begin{tabular}{c}
\includegraphics[width=0.96\textwidth]{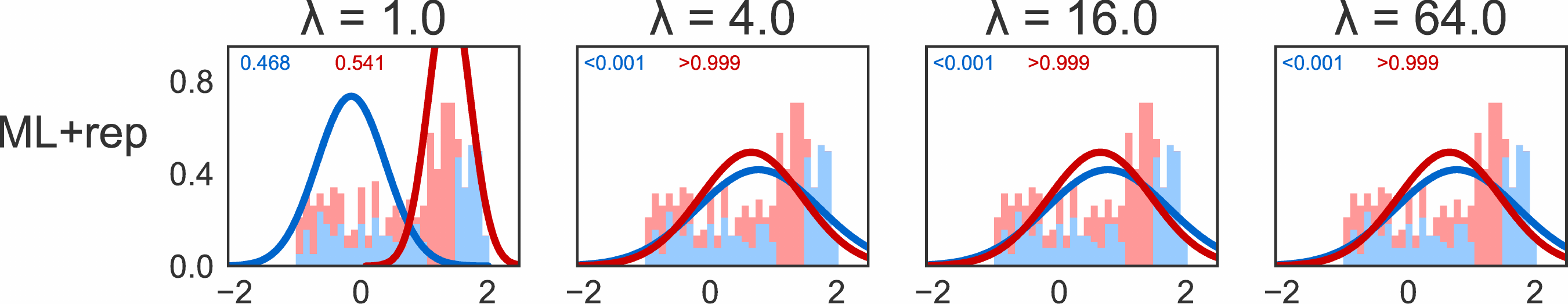}
\\
\includegraphics[width=0.96\textwidth]{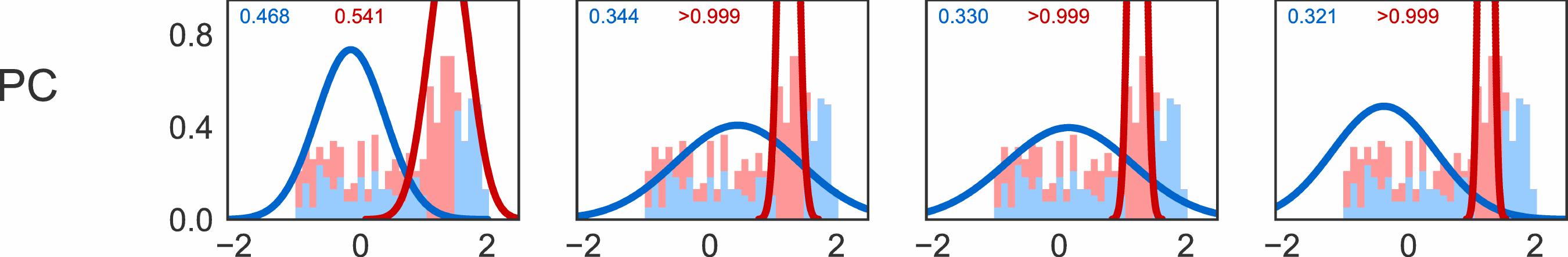}
\end{tabular}
\begin{tabular}{c c c}
\includegraphics[width=0.3\textwidth]{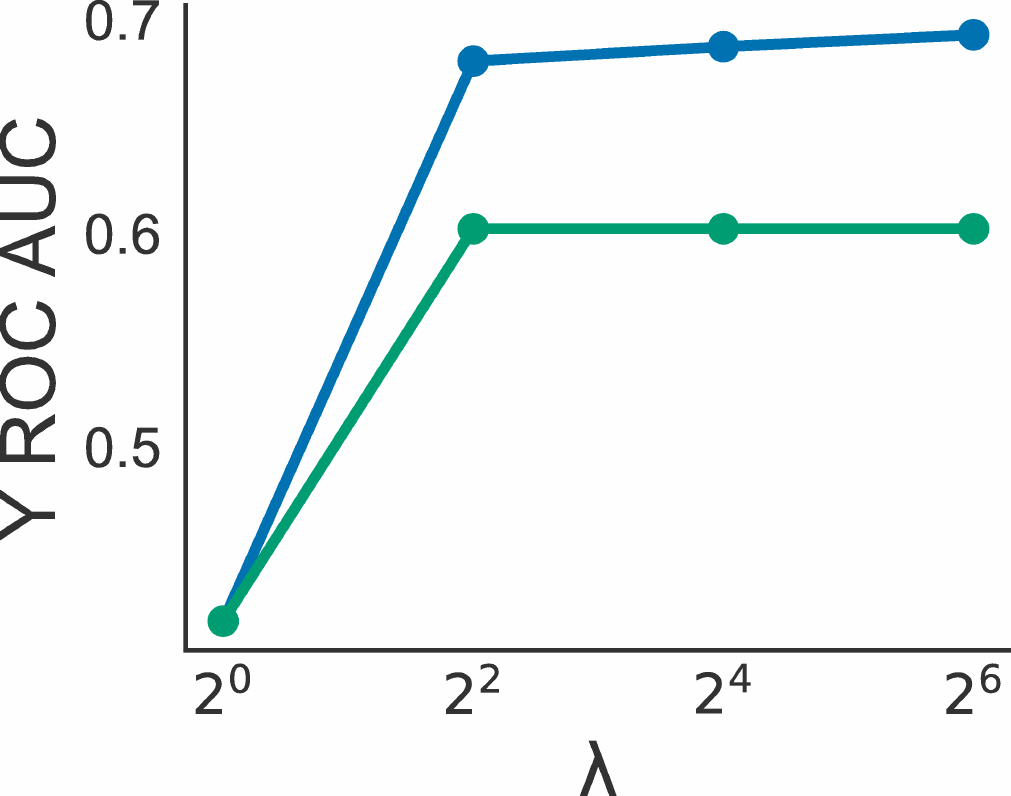}
&
\includegraphics[width=0.3\textwidth]{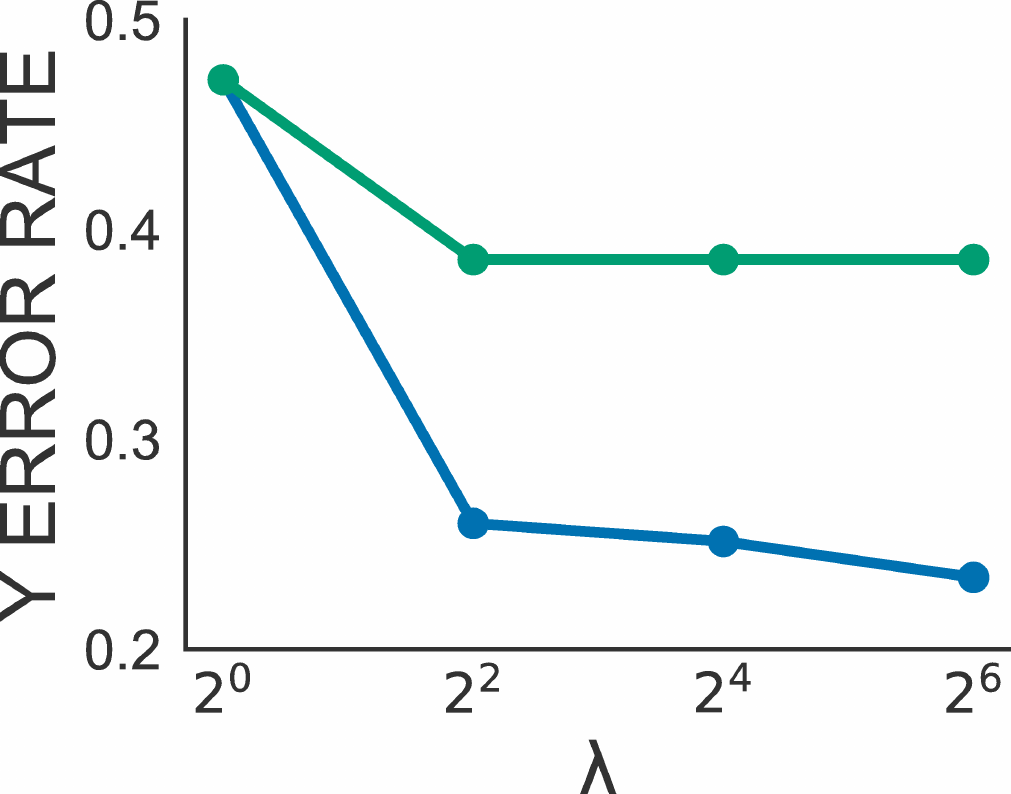}
&
\includegraphics[width=0.3\textwidth]{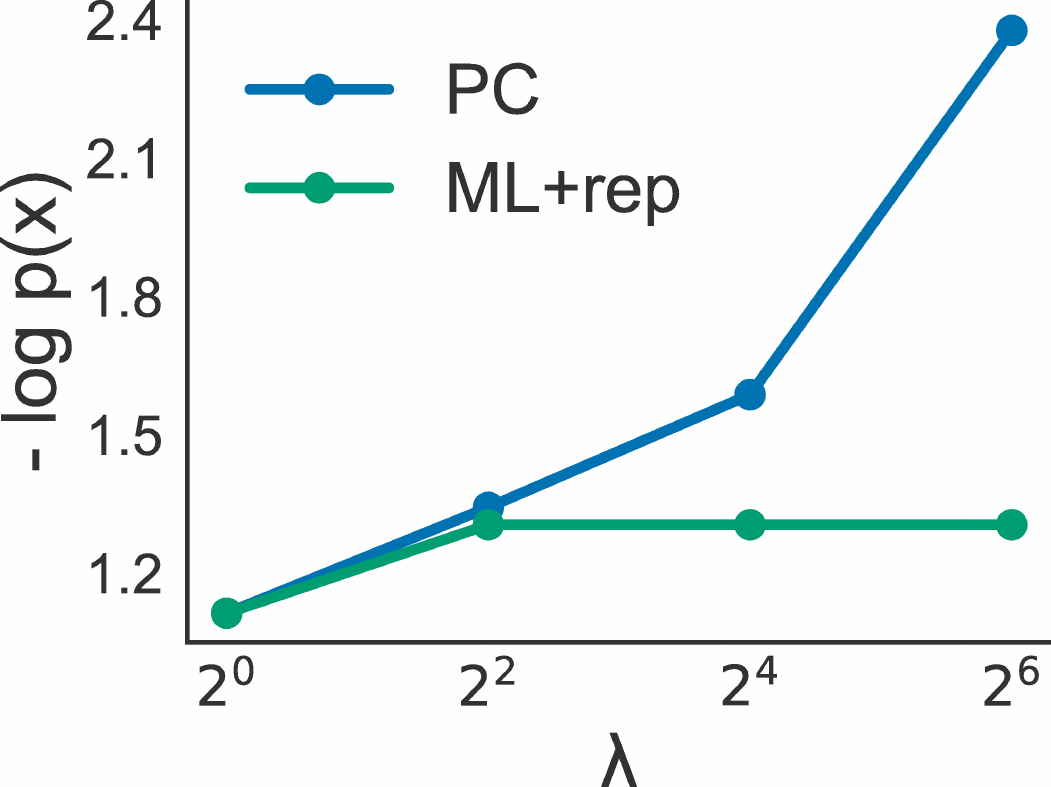}
\end{tabular}
\caption{
Toy example from Sec.~\ref{sec:mix_toy_1d_asym}: asymmetric prediction constrained (PC) training
predicts labels better than symmetric joint maximum likelihood training with label replication (ML+rep).
\emph{Top rows:}
Estimated 2-cluster Gaussian mixture model
for each training procedure under different weight values $\lambda$, taking the best of many initializations using the relevant training objective function.
Curves show the estimated 1D Gaussian distribution $\mathcal{N}(\mu_k, \sigma_k)$ for each cluster.
Upper left text in each panel gives the estimated probability $\rho_k$ that each cluster will emit a positive label.
Colors are assigned so that red cluster has higher probability of emitting positive labels.
Stacked histograms of 1-dimensional training dataset overlaid in background (blue shading means $y=0$, red means $y=1$).
\emph{Bottom row:}
Area-under-the-ROC-curve and error rate scores for predicting labels $y$ from data $x$ on \emph{training} data,  using the best solution (as ranked by each training objective) across different weight values $\lambda$.
Final panel shows negative log likelihood of data $x$ (normalized by number of data points) across same $\lambda$ values.
}
\label{fig:mix_toy_1d_asym_example}
\end{figure}

We now consider a small example to illustrate one of our fundamental contributions: that PC training is often superior to symmetric maximum likelihood training with label replication, in terms of finding models that accurately predict labels $y$ given data $x$.
We will apply supervised mixture models to a simple toy dataset with data $x_d \in \mathbb{R}$ on the real line and binary labels $y_d \in \{0, 1\}$.
The observed training dataset is shown in the top rows of Fig.~\ref{fig:mix_toy_1d_asym_example} as a stacked histogram. 
We construct the data by drawing data $x$ from three different uniform distributions over distinct intervals of the real line, which we label in order from left to right for later reference:
interval A contains 175 data points $x \in [-1, 1]$, with a roughly even distribution of positive and negative labels;
interval B contains 100 points  $x \in [1, 1.5]$ with purely positive labels;
interval C contains 75 points $x \in [1.5, 2.0]$ with purely negative labels.
Stacked histograms of the data distribution, colored by the assigned label, can be found in Fig.~\ref{fig:mix_toy_1d_asym_example}.

We now wish to train a supervised mixture model for this 
dataset. To fully specify the model, we must define concrete densities and parameter spaces.
For the data likelihood $f$, we use a 1D Gaussian distribution $\mathcal{N}(\mu_k, \sigma_k)$, with two parameters $\xi^x_k = \{\mu_k, \sigma_k \}$ for each cluster $k$.
The mean parameter $\mu_k \in \mathbb{R}$ can take any real value, while the standard deviation is positive with a small minimum value to avoid degeneracy: $\sigma_k \in (0.001, +\infty)$.
For the label likelihood $g$, we select a Bernoulli likelihood $\mbox{Bern}(\rho_k)$, 
which has one parameter per cluster: $\xi^y_k = \{\rho_k \}$, where $\rho_k \in (0, 1)$ defines the probability that labels produced by cluster $k$ will be positive.
For this example, we fix the model structure to exactly $K=2$ total clusters for simplicity.

We apply very light regularization on only the $\pi$ and $\rho$ parameters:
\begin{align}
R(\pi) = - \log \mbox{Dir}(\pi \mid 1.01, \ldots 1.01), \quad 
\ts
R(\rho) = \sum_{k=1}^K - \log \mbox{Beta}(\rho_k \mid 1.01, 1.01).
\end{align}
These choices ensure that MAP estimates of $\rho_k$ and $\pi$ are unique and always exist in numerically valid ranges (not on boundary values of exactly 0 or 1).
This is helpful for the closed-form maximization step we use for the EM algorithm for the ML+rep objective.

When using this model to explain this dataset,
there is a fundamental tension between explaining the data $x$ and the labels $y | x$: no one set of parameters $\xi$ will outrank all other parameters on both objectives. 
For example, standard joint maximum likelihood training (equivalent to our PC objective when $\lambda = 1$) happens to prefer a $K=2$ mixture model with two well-separated Gaussian clusters with means around 0 and 1.5.
This gives reasonable coverage of data density $p(x)$, but has quite poor predictive performance $p(y|x)$, because the left cluster is centered over interval A (a non-separable even mix of positive and negative examples), while the right cluster explains both B and C (which together contain 100 positive and 75 negative examples). 

Our PC training objective allows prioritizing the prediction of $y | x$ by increasing the Lagrange multiplier weight $\lambda$. Fig.~\ref{fig:mix_toy_1d_asym_example} shows that for $\lambda = 4$, the PC objective prefers the solution with one cluster (colored red) exclusively explaining interval B, which has only positive labels. The other cluster (colored blue), has wider variance to cover all remaining data points. This solution has much lower error rate ($\approx0.25$ vs. $\approx0.5$) and higher AUC values ($\approx0.69$ vs. $\approx0.5$) than the basic $\lambda = 1$ solution. Of course, the tradeoff is a visibly lower likelihood of the training data $\log p(x)$, since the higher-variance blue cluster does less well explaining the empirical distribution of $x$.
As $\lambda$ increases beyond 4, the quality of label prediction improves slightly as the decision boundaries get even sharper, but this requires the blue background cluster to drift further away from data and reduce data likelihood even more. In total, this example illustrates how PC training enables the practitioner to explore a range of possible models that tradeoff data likelihood and prediction quality.

In contrast, any amount of label replication for standard maximum likelihood training does \emph{not} reach the prediction quality obtained by our PC approach. We show trained models for replication weights values equal to 1, 4, 16, and 64 in Fig.~\ref{fig:mix_toy_1d_asym_example} (we use common notation $\lambda$ for simplicity). For all values $\lambda > 1$, we see that symmetric joint ``ML+rep'' training finds the same solution: Gaussian clusters that are exclusively dedicated to either purely positive or purely negative labels. This occurs because at training time, both $x$ and $y$ are fully observed, and thus the replicated presence of $y$ strongly cues which cluster to assign and allows completely perfect label classification. However, when we then try asymmetric prediction of $y$ given only $x$ on the same training data, we see that performance is much worse: the error rate is roughly 0.4 while our PC method achieved near 0.25. 
It is important to stress that \emph{no amount} of label replication would fix this, because the asymmetric task of predicting $y$ given only $x$ is not the focus of the symmetric joint likelihood objective.

\subsection{Toy Example: Advantage of Semisupervised PC Training}
\label{sec:mix_toy_semisuper}

\begin{figure}[!t]
\begin{minipage}{0.95\textwidth}
\begin{tabular}{c}
\includegraphics[width=\textwidth]{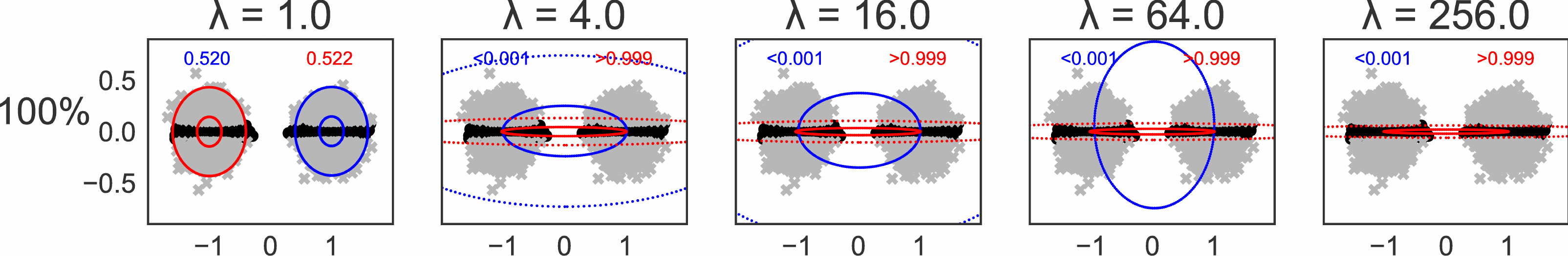}
\\
\includegraphics[width=\textwidth]{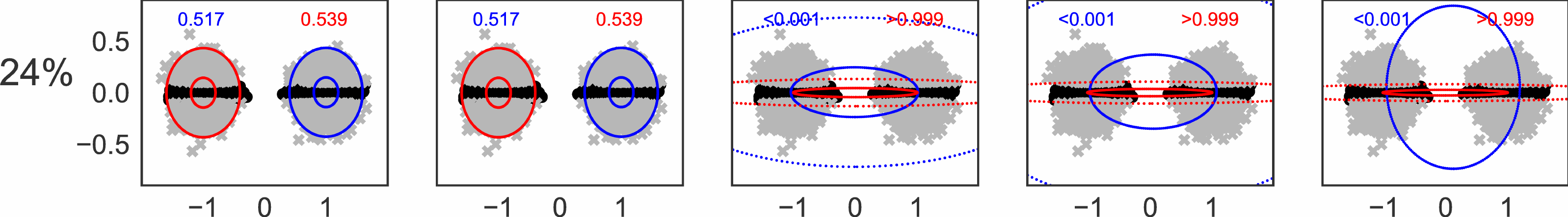}
\\
\includegraphics[width=\textwidth]{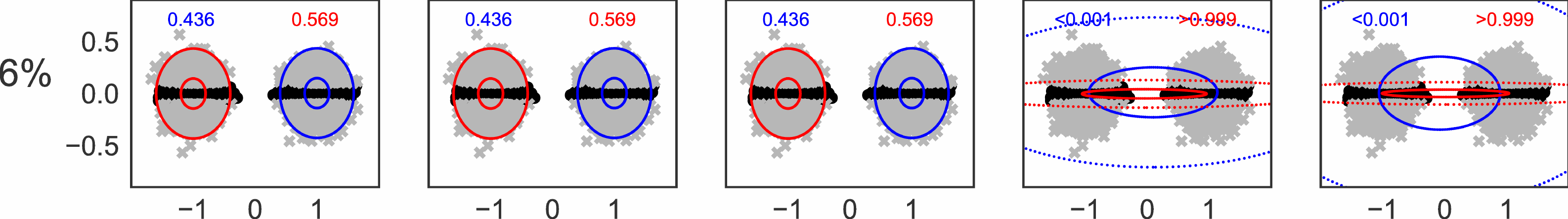}
\\
\includegraphics[width=\textwidth]{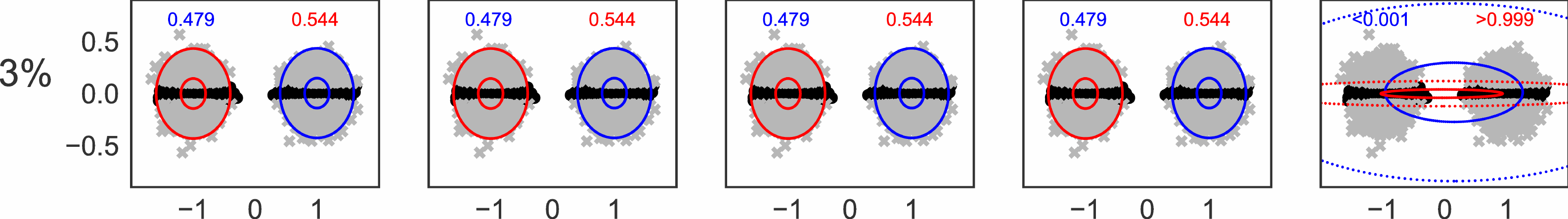}
\end{tabular}
\end{minipage}
\begin{minipage}{0.9\textwidth}
\centering 
(a) PC: Prediction-constrained
\end{minipage}
\begin{minipage}{0.95\textwidth}
\begin{tabular}{c}
~
\\
\includegraphics[width=\textwidth]{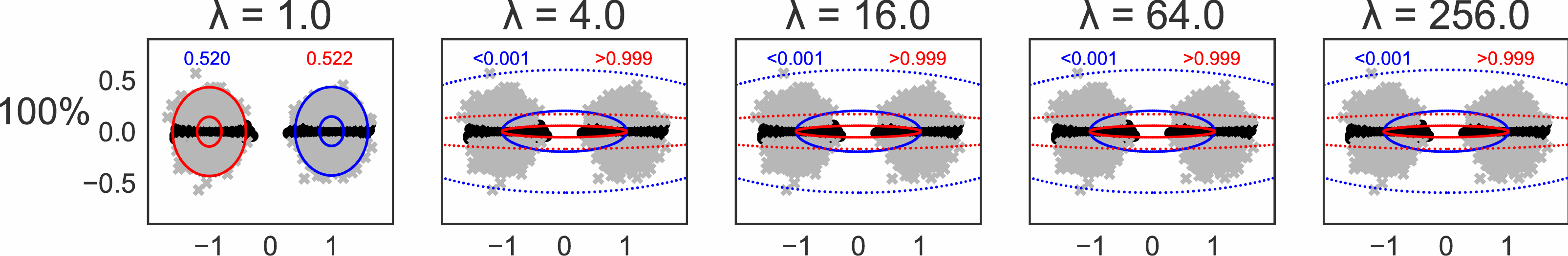}
\\
\includegraphics[width=\textwidth]{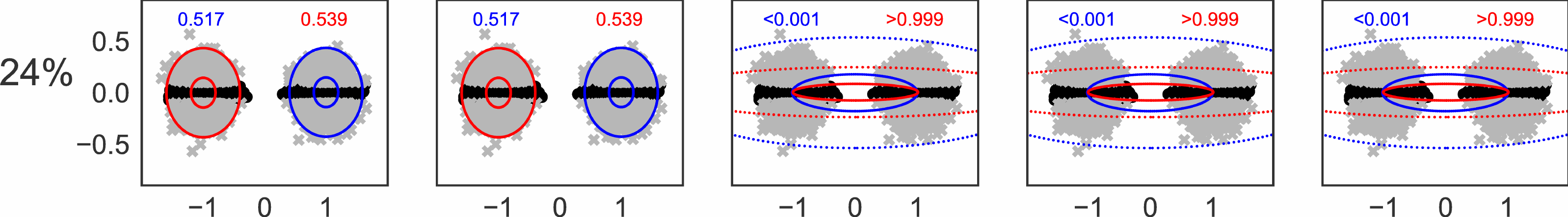}
\\
\includegraphics[width=\textwidth]{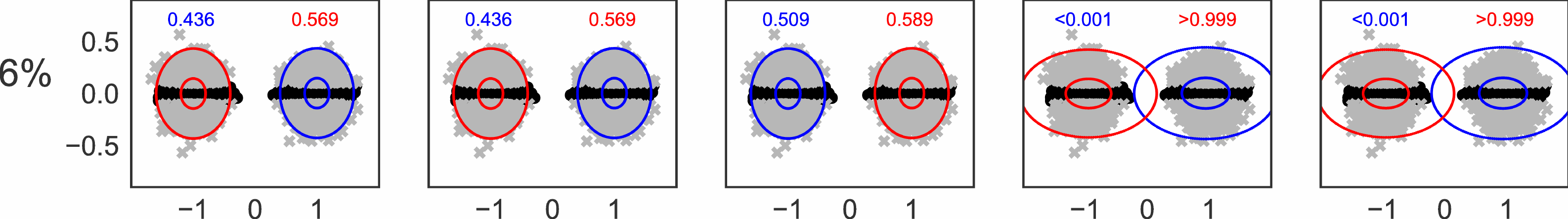}
\\
\includegraphics[width=\textwidth]{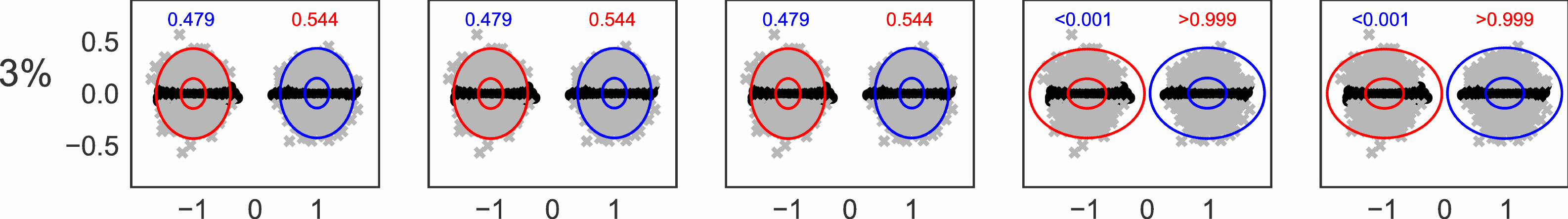}
\end{tabular}
\end{minipage}
\begin{minipage}{0.9\textwidth}
\centering
(b) ML+rep: Maximum likelihood with label replication
\end{minipage}
\caption{
Toy example from Sec.~\ref{sec:mix_toy_semisuper}: Estimated supervised mixture models produced by PC training (a) and ML+rep (b) for semi-supervised tasks with few labeled examples.
Each panel shows the 2D elliptical contours of the estimated $K=2$ cluster Gaussian mixture model
which scored best under each training objective 
using the indicated weight $\lambda$ and percentage $b$ of examples which have observed labels at training, which varies from 3\% to 100\%.
Upper text in each panel gives the estimated probability $\rho_k$ that each cluster will emit a positive label.
Colors are assigned so that red cluster has higher probability of emitting positive labels.
In the background of each panel is a scatter plot of the first two dimensions of data $x$,
with each point colored by its binary label $y$ (grey = negative, black = positive).
}
\label{fig:mix_toy_semisuper_params}
\end{figure}

\begin{figure}[!t]
\begin{tabular}{c}
\includegraphics[width=0.9\textwidth]{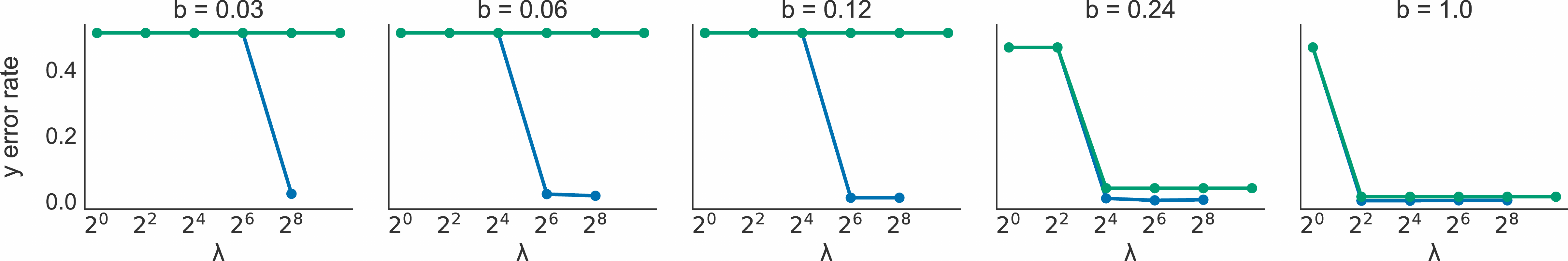}
\\
\includegraphics[width=0.9\textwidth]{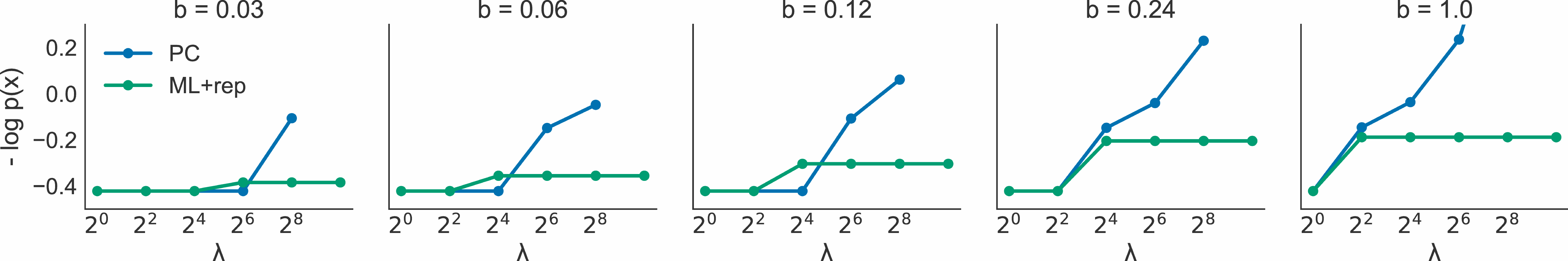}
\end{tabular}
\caption{
Toy example from Sec.~\ref{sec:mix_toy_semisuper}:
Each panel shows 
line plots of performance metrics as the PC or replication weight $\lambda$ increases, for particular percentage of data $b$ that is labeled.
Top row shows label prediction error rate (lower is better),
and bottom row shows negative data likelihood $- \log p(x)$ (lower is better).
For visualizations of corresponding parameters, see Fig.~\ref{fig:mix_toy_semisuper_params}.
}%
\label{fig:mix_toy_semisuper_lineplots}
\end{figure}

Next, we study how our PC training objective enables useful analysis of \emph{semi-supervised} datasets, which contain many unlabeled examples and few labeled examples. Again, we will illustrate clear advantages of our approach over standard maximum likelihood training in prediction quality.

The dataset is generated in two stages. First, we generate 5000 data vectors $x_d \in \mathbb{R}^5$ drawn from a mixture of 2 well-separated Gaussians with diagonal covariance matrices:
\begin{align}
\notag
x_d \sim
    \frac{1}{2} \mathcal{N}
    \Bigg(
\left[
\begin{smallmatrix}
-1 \\
0 \\
0 \\
0 \\
0
\end{smallmatrix}
\right]    
    ,
    \left[
    \begin{smallmatrix}
    2 &   &   &     &   \\
  & 1 &   &     &   \\
  &   & 1 &     &   \\
  &   &   & 0.5 &   \\
  &   &   &     & 1
    \end{smallmatrix}
    \right]
    \Bigg)
+
    \frac{1}{2} \mathcal{N}
    \Bigg(
\left[
\begin{smallmatrix}
+1 \\
0 \\
0 \\
0 \\
0
\end{smallmatrix}
\right]    
    ,
    \left[
    \begin{smallmatrix}
    2 &   &   &     &   \\
  & 1 &   &     &   \\
  &   & 1 &     &   \\
  &   &   & 1 &   \\
  &   &   &     & 0.5
    \end{smallmatrix}
    \right]
    \Bigg).
\end{align}
Next, we generate binary labels $y_d$ according to a fixed threshold rule which uses only the absolute value of the second dimension of $x_d$:
\begin{align}
y_d | x_d = \begin{cases}
    1 \quad & \mbox{~if~~} |x_{d2}| < 0.1,
 \\
    0 \quad & \mbox{~otherwise}.
 \end{cases}
\end{align}
While the full data vectors are 5-dimensional, we can visualize the first two dimensions of $x$ as a scatterplot in Fig.~\ref{fig:mix_toy_semisuper_params}. Each point is annotated by its binary label $y$: 0-labeled data points are grey 'x' markers while 1-labeled points are black 'o' markers. 
Finally, we make the problem \emph{semi-supervised} by selecting some percentage $b$ of the 5000 data points to keep labeled during training. For example if $b=50\%$, then we train using 2500 labeled pairs $\{x_d, y_d \}$ randomly selected from the full dataset as well as the remaining 2500 unlabeled data points. Our model specification is the same as the previous example: Gaussian with diagonal covariance for $f$, Bernoulli likelihood for $g$, and the same light regularization as before to allow closed-form, numerically-valid M-steps when optimizing the ML+rep objective via EM.

We have deliberately constructed this dataset so that a $K=2$ supervised mixture model is \emph{misspecified}. Either the model will do well at capturing the data density $p(x)$ by covering the two well-separated blobs with equal-covariance Gaussians, 
or it will model the predictive density $p(y | x)$ well by using a thin horizontal Gaussian to model the black $y=1$ points as well as a much larger background Gaussian to capture the rest. 
With only 2 clusters, no single model can do well at both.

Our PC approach provides a range of possible models to consider, one for each value of $\lambda$, which tradeoff these two objectives.
Line plots showing overall performance trends for data likelihood $p(x)$ and prediction quality are shown in Fig.~\ref{fig:mix_toy_semisuper_lineplots}, while the corresponding parameter visualizations are shown in Fig.~\ref{fig:mix_toy_semisuper_params}.
Overall, we see that PC training when $\lambda =1$, which is equivalent to standard ML training, yields a solution which explains the data $x$ well but is poor at label prediction. 
For all tested fractions of labeled data $b$,
as we increase $\lambda$
there exists some critical point at which this solution is no longer prefered and the objective instead favors a solution with near-zero error rate for label prediction.
For $b= 100\%$, we find a solution with near zero error rate at $\lambda = 4$, while for $b = 3\%$ we see that it takes $\lambda \gg 64$.

In contrast, when we test symmetric ML training with label replication across many replication weights $\lambda$, we see big differences between plentiful labels 
$(b \gtrapprox 20\%)$ and scarce labels $(b \lessapprox 20\%)$.
When enough labeled examples are available, high replication weights do favor the same near-zero error rate solution found by our PC approach.
However, there is some critical value of $b$ below which this solution is no longer favored, and instead the prefered solution for label replication is a pathological one: two well-separated clusters that explain the data well but have extreme label probabilities $\rho_k$.
Consider the $b=3\%, \lambda=64.0$ solution for ML+rep in Fig.~\ref{fig:mix_toy_semisuper_params}.
The red cluster explains the left blob of unlabeled data $x$ (containing about 2400 data points) as well as all positive labels $y$ observed at training, which occur in both the left and right blobs (only 150 total labels exist, of which about half are positive).
The symmetric joint ML objective weighs each data point, whether labeled or unlabeled, equally when updating the parameters $\xi^h, \xi^x$ that control $p(x)$ no matter how much replication occurs.
Thus, enough unlabeled points exert strong influence for the particular well-separated blob configuration of the data density $p(x)$, and the few labeled points can be easily explained as outliers to the two blobs.
In contrast, our PC objective by construction allows upweighting the influence of the asymmetric prediction task on \emph{all} parameters, including $\xi^h, \xi^x$. Thus, even when replication happens to yield good predictions when all labels are observed, it can yield pathologies with few labels that our PC easily avoids.

\section{Case Study: Prediction-constrained Topic Models}
\label{sec:topics}

We now present a much more thorough case-study of prediction-constrained \emph{topic
models}, building on latent Dirichlet allocation (LDA)~\citep{blei2003lda} and its downstream supervised extension sLDA~\citep{blei2007sLDA}.
The \emph{unsupervised} LDA topic model 
takes as observed data a collection of $D$ documents, or more generally, $D$ groups of
discrete data.  Each document $d$ is represented by counts of $V$
discrete word types or features, $x_d \in \mathbb{Z}_+^V$. 
We explain these observations via $K$ latent clusters or \emph{topics},
such that each document exhibits \emph{mixed-membership} across these topics. Specifically, in terms of our general downstream LVM model family the model assumes a hidden variable $\pi_d \in h_d$ such that $\pi_d = [ \pi_{d1} \ldots \pi_{dK} ]$ is a vector of $K$ positive numbers that sum to one, indicating which fraction of the document is explained by each topic $k$. The generative model is:
\begin{align}
P:\quad \pi_d | \alpha &\sim \mbox{Dir}( \pi_d \mid \alpha), \nonumber
\\
F:\quad x_d | \pi_d, \phi &\sim \mbox{Mult}( x_d \mid 
        \textstyle \sum_{k=1}^K \pi_{dk} \phi_{k}, N_d ).
\label{eq:lda_generative_model}        
\end{align}
Here, the hidden variable prior density $P$ is chosen to be a symmetric Dirichlet with parameters $\xi^h = \{ \alpha \}$, where $\alpha > 0$ is a scalar. Similarly, the data likelihood parameters are defined as $\xi^x = \{\phi_k\}_{k=1}^K$, where each topic $k$ has a parameter vector $\phi_{k}$ of $V$ positive numbers (one for each vocabulary term) that sums to one. The value $\phi_{kv}$ defines the probability of generating word $v$ under topic $k$.
Finally, we assume that the size of document~$d$ is observed as $N_d$.

In the supervised setting, we assume that each document $d$ also has an observed target value $y_d$. For our applications, we'll assume this is one or more binary labels, so 
$y_d \in \{0,1\}$, but we emphasize other types of $y$ values are easily possible via generalized linear models~\citep{blei2007sLDA}.
Standard supervised topic models like sLDA assume labels and
word counts are conditionally independent given topic probabilities
$\pi_d$, via the label likelihood:
\begin{align}
G: y_d | \pi_d, \eta &\sim \mbox{Bern}( y_d \mid \sigma(
        \textstyle \sum_{k=1}^K \pi_{dk} \eta_{k} ) ),
\label{eq:target_model}        
\end{align}
where $\sigma(x) = (1+e^{-x})^{-1}$ is the logit function, and $\eta \in \mathbb{R}^K$ is a vector of real-valued regression parameters.
Under this model, large positive values $\eta_k \gg 0$ imply that high usage of topic $k$ in a given document (larger $\pi_{dk}$) will lead to predictions of a positive label $y_d = 1$. Large negative values $\eta_k \ll 0$ imply high topic usage leads to a negative label prediction $y_d = 0$.

The original sLDA model~\citep{blei2007sLDA} represents the count likelihood via $N_d$ independent assignments $z_{dn} \sim \mbox{Cat}(\pi_d)$ of word tokens to topics, and generates labels 
$y_d \sim \mbox{Bern}(y_d \mid \sigma(\sum_{k=1}^K \bar{z}_{dk} \eta_k))$,
where $\bar{z}_d$ is a vector on the $K-$dimensional probability simplex given the empirical distribution of the token-to-topic assignments: $\bar{z}_{dk} \triangleq N_d^{-1}\sum_{n} \delta_{k}(z_{dn})$ and $E[\bar{z}_{dk}] = \pi_{dk}$.
To enable more efficient inference algorithms, we analytically marginalize these topic assignments away in Eq.~(\ref{eq:lda_generative_model},\ref{eq:target_model}).

\paragraph{PC objective for sLDA.}
Applying the PC objective of
Eq.~\eqref{eq:pc_objective_unconstrained_format} to the sLDA model gives:
\begin{align}
\underset{\phi,\eta,\alpha}{\min}
     - \sum_{d=1}^D \log p(x_d \mid \phi, \alpha ) - 
          \lambda \sum_{d=1}^D \log p( y_d \mid x_d , \phi , \eta , \alpha )
       + R( \alpha, \phi, \eta ).
\label{eq:slda_pc_objective_unconstrained_intractable}
\end{align}
Computing $p(x_d \mid \phi, \alpha )$ and $p( y_d \mid x_d , \phi , \eta , \alpha )$ involves marginalizing out the latent variables $\pi_d$:
\begin{align}
p(x_d \mid \phi, \alpha) &=
    \int_{\Delta^K}
    \mbox{Mult}(x_d \mid 
        \textstyle \sum_{k=1}^K \pi_{dk}\phi_k)
    \mbox{Dir}(\pi_d \mid \alpha)
    \;d\pi_d,\\
    p( y_d \mid x_d , \phi , \eta , \alpha ) &= \int_{\Delta^K}
    \mbox{Bern}( y_d \mid
        \sigma( \textstyle \sum_{k=1}^K \pi_{dk}\eta_k) )
        p( \pi_d \mid x_d , \phi , \alpha )
    \;d\pi_d.
\label{eq:slda_marginals}
\end{align}
Unfortunately, these integrals are intractable. To gain traction, we first contemplate an objective that \emph{instantiates} $\pi_d$ rather than marginalizes $\pi_d$ away:
\begin{align}
\min_{\pi, \phi, \eta, \alpha}  \Big[
\sum_{d=1}^D  - \log p(\pi_d | \alpha) - \log f(x_d \mid \pi_d, \phi) - \lambda \log g(y_d \mid \pi_d, \eta)
\Big] 
+ R(\phi, \eta, \alpha).
\label{eq:slda_mlrep_objective}
\end{align}
However, this objective is simply a version of maximum likelihood with label-replication from Sec.~\ref{sec:problems}, albeit with hidden variables instantiated rather than marginalized. The same poor prediction quality issues will arise due to its inherent symmetry.
Instead, because we wish to train under the same assymetric conditions needed at test time, where we have $x_d$ but not $y_d$, we do not instantiate $\pi_d$ as a free variable but \emph{fix} $\pi_d$ to a deterministic mapping of the words $x_d$ to the topic simplex. 
Specifically, we fix to the maximum a-posteriori (MAP) solution
$\pi_d = \mbox{argmax}_{\pi \in \Delta^K} \log p(\pi_d | x_d, \alpha)$, which we write as a deterministic function:
$\pi_d \gets \mbox{MAP}( x_d , \phi , \alpha )$. We show in Sec.~\ref{sec:inference} that this deterministic embedding of any document's data $x_d$ onto the topic simplex is easy to compute.
Our chosen embedding can be seen as a feasible
approximation to the full posterior $p( \pi_d | x_d , \phi , \alpha )$ needed in Eq.~\eqref{eq:slda_marginals}.
This choice which respects the need to use the same embedding of observed words $x_d$ into low-dimensional $\pi_d$ in both training and test scenarios.

We can now write a tractable training objective we wish to minimize:
\begin{align}
\label{eq:slda_pc_objective_differentiable}
\mathcal{J}(\phi,\eta,\alpha) &=
    -     \Big[ 
    \sum_{d=1}^D 
    \log p(\mbox{MAP}( x_d , \phi , \alpha ) \mid \alpha)
    + \log f(x_d \mid \mbox{MAP}( x_d , \phi , \alpha ) , \phi )
    \Big]
    \\ \notag 
    & \quad
    - \lambda \Big[ 
        \sum_{d=1}^D
            \log g( y_d \mid \mbox{MAP}( x_d , \phi , \alpha ), \eta )
        \Big]
    \Big]
       + R( \phi, \eta, \alpha ).
\end{align}
This objective is both tractable to evaluate and fixes the asymmetry issue
of standard sLDA training, because the model is forced to learn the embedding function which will be used at test time.

\paragraph{Previous training objectives for sLDA.}
Originally, the sLDA model was trained via a variational EM algorithm that optimizes a lower bound on the marginal likelihood of the observed words and labels~\citep{blei2007sLDA}; MCMC sampling for posterior estimation is also possible. This treatment ignores the cardinality mismatch and assymetry issues, making it difficult to make good predictions of $y$ given $x$ under conditions of model mismatch.
Alternatives like MED-LDA~\citep{zhu2012medlda} offered alternative objectives which try to enforce constraints on the loss function given expectations under the approximate posterior, yet this objective still ignores the crucial asymmetry issue. We also showed earlier in Sec.~\ref{sec:problems} that some MED objectives can be reduced to ineffective maximum likelihood with label-replication.

Recently, \citet{chen2015bplda} developed \emph{backpropagation} methods called BP-LDA and BP-sLDA for the unsupervised and supervised versions of LDA. They train using extreme cases of our end-to-end weighted objective in Eq.~\eqref{eq:slda_pc_objective_differentiable},
where for supervised BP-sLDA the entire data likelihood term $\log p(x_d \mid \pi_d)$ is omitted completely,
and for unsupervised BP-LDA the entire label likelihood $\log p(y_d \mid \pi_d)$ is omitted.
In contrast, 
our overriding goal of
guaranteeing some minimum prediction quality via our PC objective in
Eq.~\eqref{eq:pc_objective_unconstrained_format} leads to a Lagrange multiplier
$0 < \lambda < \infty$ which allows us to systematically balance the
generative and discriminative objectives.
BP-sLDA offers no such tradeoff, and we will see in later experiments that while its label predictions are sometimes good, the underlying topic model is quite terrible at explaining heldout data and yields difficult-to-interpret topic-word distributions.

\subsection{Inference and learning for
Prediction-Constrained LDA}
\label{sec:inference}
Fitting the sLDA model to a given dataset using our PC optimization objective in Eq.~\eqref{eq:slda_pc_objective_differentiable} requires two concrete procedures: per-document inference to compute the hidden variable $\pi_d$, and global parameter estimation of the topic-word parameters $\phi$ and logistic regression weight vector $\eta$.
First, we show how the MAP embedding $\pi_d \gets \mbox{MAP}(x_d, \phi, \alpha)$ can be computed via several iterations of an exponentiated gradient procedure with convex structure.
Second, we show how we can differentiate through the entire objective to perform gradient descent on our parameters of interest $\phi$ and $\eta$.
While in our experiments, we assume that the prior concentration parameter $\alpha > 0$ is a fixed constant, this could easily be optimized as well via the same procedure.

\paragraph{MAP inference via exponentiated gradient iterations.}
\citet{sontag2011complexityoflda} define the document-topic MAP estimation problem for LDA as:\begin{align}
\pi_d' = \max_{\pi_d \in \Delta^{K}} \ell(\pi_d, x_d, \phi, \alpha),
\quad \ell(\pi_d, x_d, \phi, \alpha) = \log \mbox{Mult}(x_d \mid \pi_d^T \phi) + \log \mbox{Dir}(\pi_d \mid \alpha).
\label{eq:objective_for_pi_d}
\end{align}
This problem is convex for $\alpha \geq 1$ and non-convex otherwise. For the convex case, they suggest an iterative \emph{exponentiated gradient} algorithm \citep{kivinen1997exponentiated_gradient}. This procedure begins with a uniform probability vector, and iteratively performs elementwise multiplication with the exponentiated gradient until convergence using a scalar stepsize $\nu > 0$:
\begin{align}
\mbox{init:~~}
\pi^{0}_{d} \gets \Big[\frac{1}{K} \ldots \frac{1}{K}\Big],
\quad
\mbox{repeat:~}
\pi^{t}_{dk} \gets \frac{p^t_{dk}}{ \sum_{j=1}^{K} p^t_{dj} },
\quad
p^t_{dk} = \pi^{t-1}_{dk} \cdot e^{ \nu \nabla \ell(\pi^{t-1}_{dk})}.
\end{align}
With small enough steps, the final result after $T$ iterations converges
to the MAP solution.
We thus define our embedding function $\pi_d \gets \mbox{MAP}(x_d, \phi, \alpha)$
to be the outcome of $T$ iterations of the above procedure. We find $T \approx 100$ iterations and a step size of $\nu \approx 0.005$ work well. Line search for $\nu$ could reduce the number of iterations needed (though increase per-iteration costs).

Importantly, \citet{taddy2012topicmodelmapestimation} points out that while the general non-convex case $\alpha < 1$ has no single MAP solution for $\pi_d$ in the simplex due to the multimodal sparsity-promoting Dirichlet prior, a simple reparameterization into the softmax basis \citep{mackay1997ensemble} leads to a unimodal posterior and thus a unique MAP in this reparameterized space. Elegantly, this softmax basis solution for a particular $\alpha < 1$ has the same MAP estimate as the simplex MAP estimate for the ``add one'' posterior: $p( \pi_d | x_d , \phi, \alpha + 1)$. Thus, we can use our exponentiated gradient procedure to reliably perform \emph{natural parameter} MAP estimation even for $\alpha < 1$ via this ``add one'' trick.

\paragraph{Global parameter estimation via stochastic gradient descent.}

To optimize the objective in Eq.~\eqref{eq:slda_pc_objective_differentiable}, we realize first that the iterative MAP estimation function above is fully \emph{differentiable} with respect to the parameters $\phi, \eta,$ and $\alpha$, as are the probability density functions $p, f,$, and $g$.
This means the \emph{entire} objective $\mathcal{J}$ is differentiable and modern gradient descent methods may be applied easily. 
Of course, this requires standard transformations of constrained parameters like the topic-word distributions $\phi$ from the simplex to unrestricted real vectors.
Once the loss function is specified via unconstrained parameters, we  perform automatic differentiation to compute gradients and then perform gradient descent via the Adam algorithm \citep{kingma2014adam}, which easily allows stochastically sampling minibatches of data for each gradient update.
In practice, we have developed Python implementations based on both Autograd \citep{maclaurin2015autograd} and Tensorflow \citep{tensorflow2015whitepaper}, which we plan to release to the public.

Earlier work by \citet{chen2015bplda} optimized their fully discriminative objective via a mirror descent algorithm directly in the constrained parameters $\phi$, using manually-derived gradient computations within a heroically complex implementation in the C\# language. Our approach has the advantage of easily extending to other supervised loss functions without need to derive and implement gradient calculations, although the automatic differentation can be slow.

\paragraph{Hyperparameter selection.}
The key hyperparameter of our prediction-constrained LDA algorithm is the Lagrange multiplier $\lambda$.  Generally, for topic models of text data $\lambda$ needs to be on the order of the number of tokens in the average document, though it may need to be much larger depending on how much tension exists between the unsupervised and supervised terms of the objective. If possible, we suggest trying a range of logarithmically spaced values and selecting the best on validation data, although this requires expensive retraining at each value. This can be somewhat mitigated by using the final parameters at one $\lambda$ value as the initial parameters at the next $\lambda$ value, although this may not escape to new preferred basins of attraction in the overall non-convex objective.

\subsection{Supervised LDA Experimental Results}
\label{sec:slda_results}

We now assess how well our proposed PC training of sLDA, which we hereafter abbreviate as \emph{PC-LDA}, achieves its \emph{simultaneous} goals of solid heldout prediction of labels $y$ given $x$ while maintaining reasonably interpretable explanations of words $x$. We test the first goal by comparing to discriminative methods like logistic regression and supervised topic models, and the latter by comparing to unsupervised topic models. For full descriptions of all datasets and protocols, as well as more results, please see the appendix.

\paragraph{Baselines.} Our discriminative baselines include logistic regression (with a validation-tuned $L_2$ regularizer), the fully supervised BP-sLDA algorithm of \citet{chen2015bplda}, 
and the supervised MED-LDA Gibbs sampler~\citep{zhu2013gibbsmaxmargin}, which should improve on the earlier variational methods of the original MED-LDA variational algorithm in~\citet{zhu2012medlda}.
We also consider own implementation of standard coordinate-ascent variational inference for both unsupervised (VB LDA) and supervised (VB sLDA) topic models. Finally, we consider a vanilla Gibbs sampler for LDA, using the Mallet toolbox~\citep{MALLET}. We use third-party public code when possible for single-label-per-document experiments, but only our own PC-LDA and VB implementations support multiple binary labels per document, which occur in our later Yelp review label prediction and electronic health record drug prediction tasks. For these datasets only, the method we call BP-sLDA is a special case of our own PCLDA implementation (removing the data likelihood term), which we have verified is comparable to the single-target-only public implementation but allows multiple binary targets.

\paragraph{Protocol.} For each dataset, we reserve two distinct subsets of documents: one for validation of key hyperparameters, and another to report heldout metrics. All topic models 
are run from multiple random initializations of $\phi, \eta$ (for fairness, all methods use same set of predefined initializations of these parameters).
We record point estimates of topic-word parameters $\phi$ and logistic regression weights $\eta$ at defined intervals throughout training, and we select the best pair on the validation set (early stopping). 
For Bayesian methods like GibbsLDA, we select Dirichlet concentration hyperparameters $\alpha, \tau$ via a small grid search on validation data, while for PC-LDA and BP-sLDA we set $\alpha = 1.001, \tau = 1.001$ as recommended by \citet{chen2015bplda}.
For all methods, given a snapshot of parameters $\phi, \eta$, we evaluate prediction quality via area-under-the-ROC-curve (AUC) and 0/1 error rates using the prediction rule $\mbox{Pr}(y_d = 1) = \sigma(\eta^T \mbox{MAP}(x_d, \phi, \alpha))$. We evaluate data model quality by computing a variational bound on heldout per-token log perplexity: $(\sum_d N_d)^{-1} \sum_{d=1}^D \log p(x_d | \alpha, \phi)$, where $N_d$ counts tokens in document $d$.

\begin{figure}[t!]
\includegraphics[width=\textwidth]{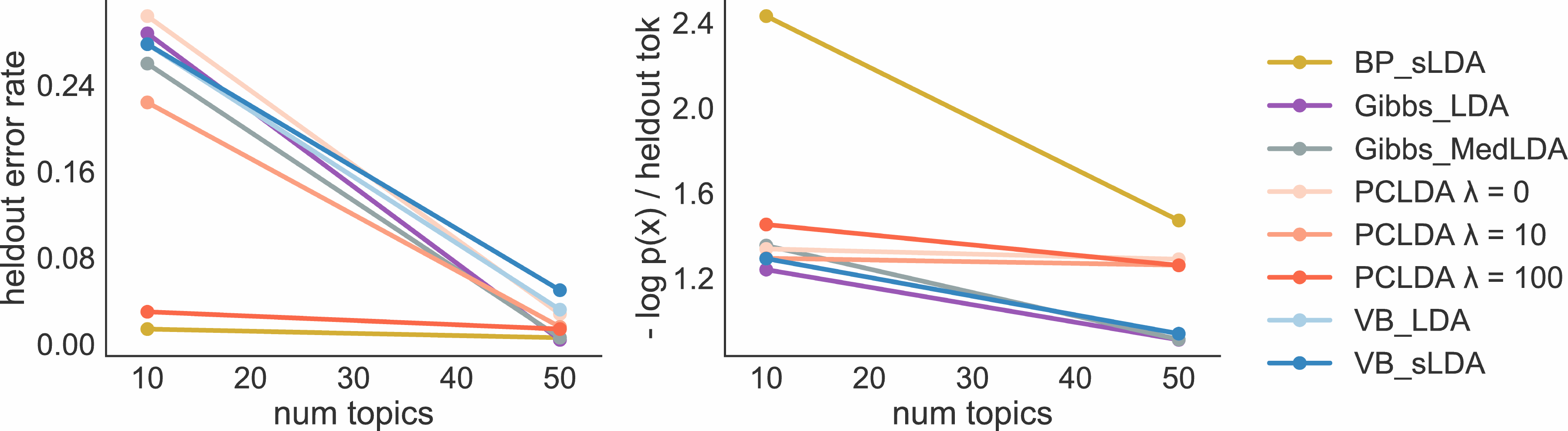}
\caption{
Vowels-from-consonants task: Heldout prediction error rates (left, lower is better) and negative log data probabilities (right, lower is better).
With enough topics ($K>30$), good unsupervised topic models can classify very well. However, for low numbers of topics ($K=10$), because consonants outnumber vowels many methods try to explain these better than the vowels.
Only PCLDA and BP sLDA are good at cleanly separating the 5 vowels with low capacity models, and PC LDA offers 
much better heldout $x$ data predictions than BP sLDA (which does not optimize $p(x)$ at all).
}%
\label{fig:toy_letters_heldout_metrics}
\end{figure}

\begin{figure}[t!]
\begin{tabular}{c}
\begin{minipage}{3cm}
example docs
\end{minipage}%
\begin{minipage}{0.75\textwidth}
\includegraphics[width=\textwidth]{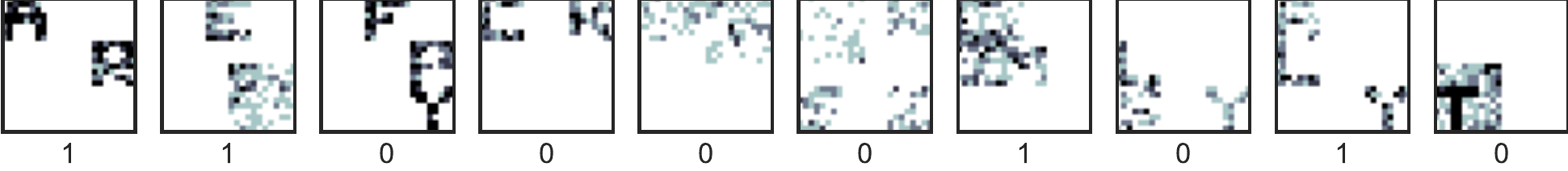} 
\end{minipage}
\\
\begin{minipage}{3cm}
true topics (10 of 30)
\end{minipage}%
\begin{minipage}{0.75\textwidth}
\includegraphics[width=\textwidth]{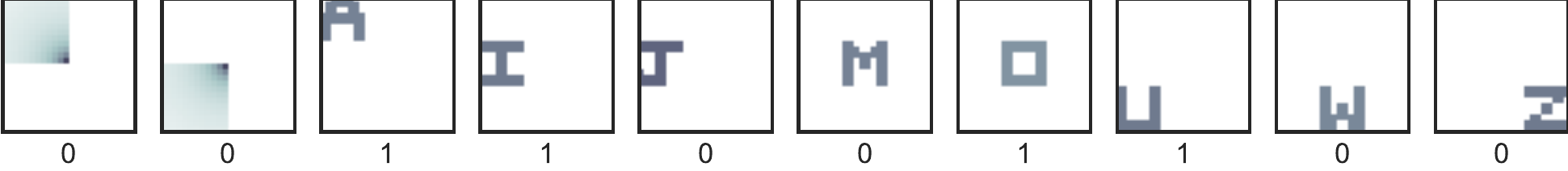} 
\end{minipage}
\\
\begin{minipage}{3cm}
0.29 ~\small{Gibbs LDA}
\end{minipage}%
\begin{minipage}{0.75\textwidth}
\includegraphics[width=\textwidth]{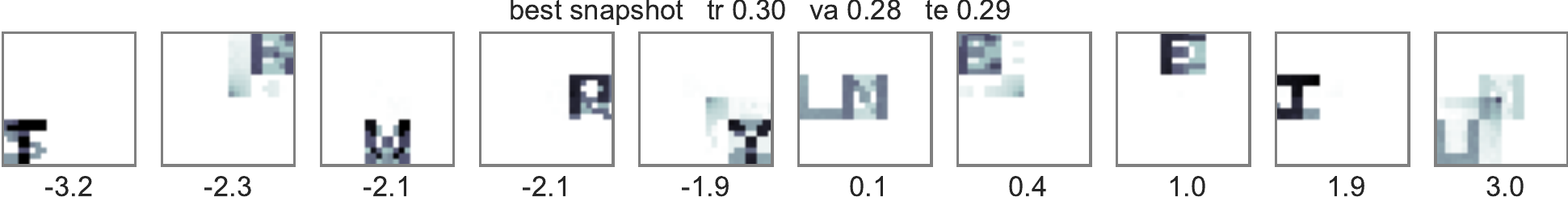} 
\end{minipage}
\\
\begin{minipage}{3cm}
0.22 ~\small{PCLDA $\lambda$=10}
\end{minipage}%
\begin{minipage}{0.75\textwidth}
\includegraphics[width=\textwidth]{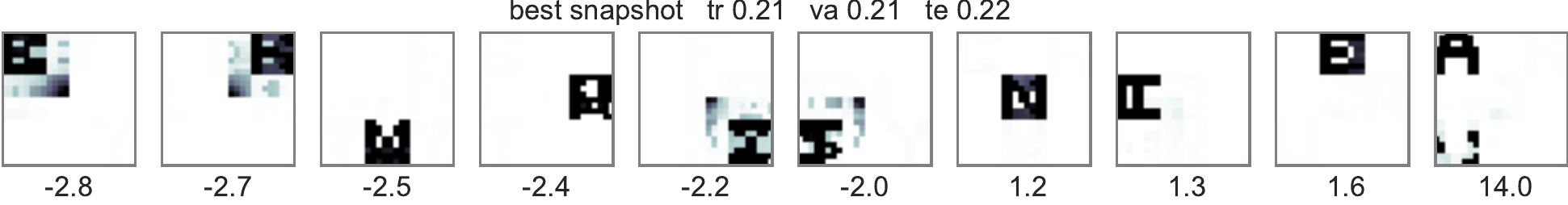} 
\end{minipage}
\\
\begin{minipage}{3cm}
0.03 ~\small{\textbf{PCLDA} $\lambda$=100}
\end{minipage}%
\begin{minipage}{0.75\textwidth}
\includegraphics[width=\textwidth]{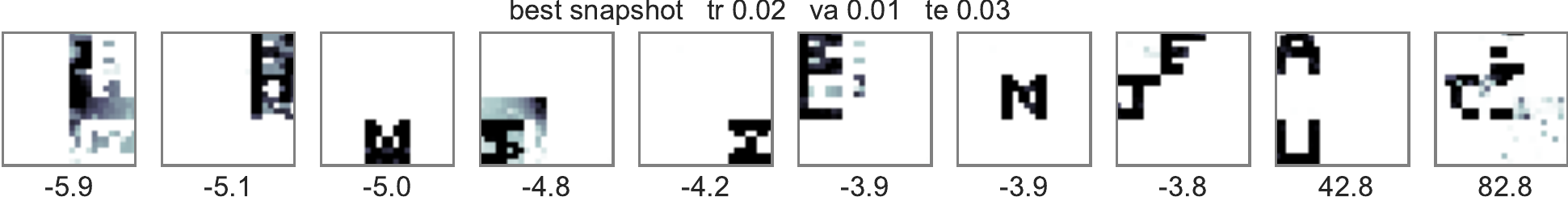} 
\end{minipage}
\\
\begin{minipage}{3cm}
0.01 ~\small{BP sLDA}
\end{minipage}%
\begin{minipage}{0.75\textwidth}
\includegraphics[width=\textwidth]{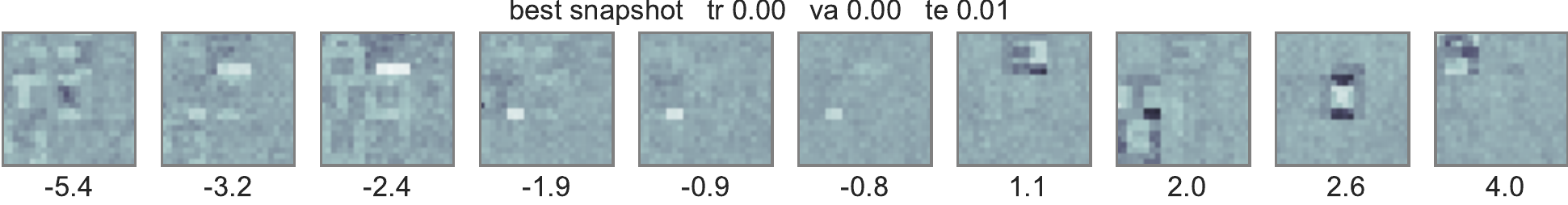}
\end{minipage}
\\
\begin{minipage}{3cm}
0.20 ~\small{MedLDA}
\end{minipage}%
\begin{minipage}{0.75\textwidth}
\includegraphics[width=\textwidth]{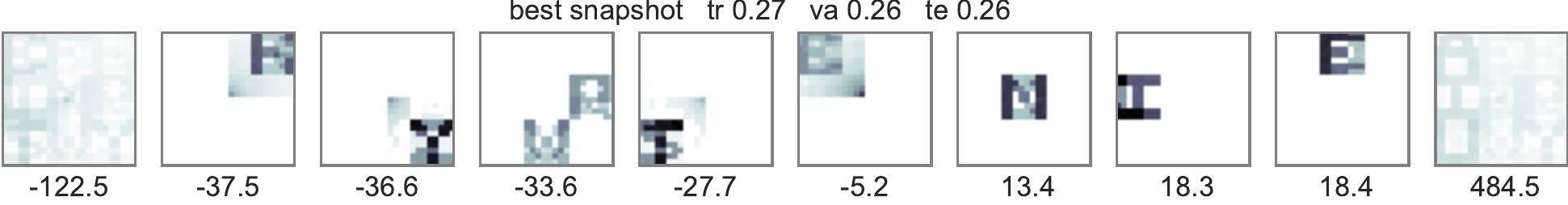}
\end{minipage}
\end{tabular}
\caption{
Vowels-from-consonants task:
Rows 1-2: example documents and true generative topics for this task.
Rows 3-end: 
Heldout error rates (left) and learned topic-word parameters for best $K=10$ model from each method.
Unsupervised Gibbs LDA, Supervised MedLDA, and PCLDA with low weight ($\lambda = 10$) have high error rates, indicating little influence of supervision into the task.
BP sLDA achieves very low error rate at the expense of messy topic-word parameters not tuned to predict $p(x)$.
However, PCLDA $\lambda=100$ reaches similar error rates while having more interpretable topics that separate E from F and I from J.
}
\label{fig:toy_letters_large}
\end{figure}

\paragraph{Tasks.}
We apply our PC-LDA approach to the following tasks: 
\begin{itemize}[leftmargin=*]
\item \textbf{Vowels-from-consonants.}
To study tradeoffs between models of $p(x)$ and $p(y|x)$, 
we built a toy ``vowels-from-consonants'' task, where each document 
$x_d$ is a sparse count vector of pixels in square grid, illustrated in Fig.~\ref{fig:toy_letters_large}. Data is generated from an LDA model with 30 total topics: 26 ``letter'' topics as well as 4 more common ``background'' topics.
Documents are labeled $y_d=1$ only if at least one vowel  (A, E, I, O, or U) appears. 
Several letters are easily confused: the pairs (E, F), (I, J) and (O, M) share many of the same pixels.
By design, even unsupervised LDA with $K>30$ does well here, but the regime of $K < 30$ assesses how well supervised methods use label cues to form topics for the targeted vowels  rather than the more plentiful consonants.

\item \textbf{Movie and Yelp reviews.}
Our movie review task \citep{pang2005moviereviews} contains 5005 documents, with documents $x_d$ drawn from the published reviews of professional movie critics. Each document has one binary label $y_d$ (1 means above average rating, 0 otherwise).
Our Yelp task~\citep{yelpdataset2016}
contains 23159 documents, each aggregating all reviews about a single restaurant into one bag of words $x_d$ with 10,000 possible vocabulary terms. Each document also has 7 possible binary attributes (takes-reservations, offers-delivery, offers-alcohol, good-for-kids, expensive-price, has-outdoor-patio, and offers-wifi).

\item \textbf{Predicting successful antidepressants from health records.} Finally, we consider predicting which subset of 10 common antidepressants will be successful for a patient with major depressive disorder given a sparse bag-of-codewords summary of the electronic health record (EHR).  These are real deidentified data from 64431 patients at tertiary care hospital and related outpatient centers.  Table~\ref{table:psych_results} contains results for label prediction, and Fig.~\ref{fig:psych_topics} visualizes top word lists from learned topics.
\end{itemize}

\begin{figure}
\includegraphics[width=\textwidth]{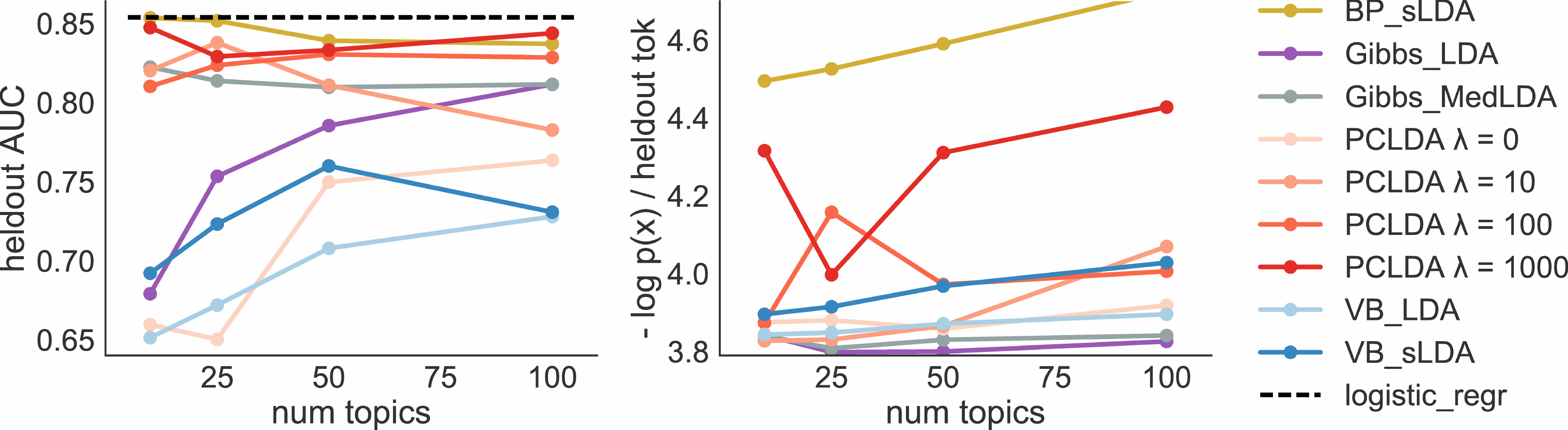}
\caption{
Movie reviews task:
Area-under-ROC curve for binary sentiment prediction (left, higher is better) and negative heldout log probability of tokens (right, lower is better).
Our PCLDA makes competitive label predictions (left) while maintaining better data models than BPsLDA (right).
}%
\label{fig:movie_review_heldout_metrics}
\end{figure}

\begin{figure}[t!]
\includegraphics[width=\textwidth]{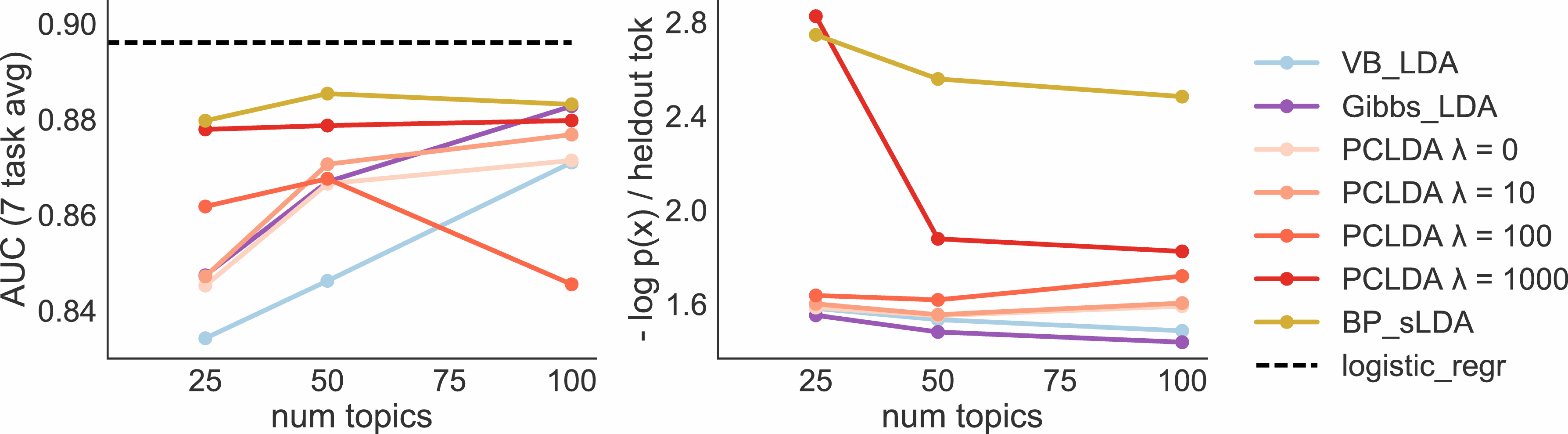}
\caption{
Yelp reviews task:
Area-under-ROC curve for label prediction (left, higher is better) and negative heldout log probability of tokens (right, lower is better).
Here, we report the \emph{average} AUC across the 7 possible review labels in Table~\ref{table:yelp_results}.
}%
\label{fig:yelp}
\end{figure}

\paragraph{Prediction-constrained LDA can match discriminative baselines like logistic regression when predicting labels.}
Fig.~\ref{fig:movie_review_heldout_metrics} shows that PC-LDA with high $\lambda$ values is competitive with logistic regression as well as BPsLDA and MED-LDA. The AUC numbers in Table \ref{table:psych_results} show that our method is at worst within 0.03 of competitor AUC scores, and in some cases (paroxetine, venlafaxine, amitriptyline) better than logistic regression. In the Yelp task in Fig.~\ref{fig:yelp}, our PC-LDA $\lambda=1000$ (dark red line) achieves similar AUC numbers to the purely discriminative BP-sLDA (gold line), and only slightly worse performance than logistic regression.

\begin{table}[t!]
\begin{tabular}{c | r r r r r}
\toprule
task & Gibbs LDA & PCLDA $\lambda$ = 10 
& PCLDA $\lambda$ = 1000 
& BP sLDA 
& logistic regr
\\
~&
K=50 &
K=50 &
K=50 &
K=50 &
~
\\
\midrule
reservations & 0.921 & 0.920 & 0.934 & 0.934 & 0.934
\\
delivery & 0.870 & 0.853 & 0.873 & 0.873 & 0.886
\\
alcohol & 0.929 & 0.928 & 0.948 & 0.950 & 0.952
\\
kid friendly & 0.889 & 0.901 & 0.899 & 0.908 & 0.919
\\
expensive & 0.921 & 0.919 & 0.929 & 0.935 & 0.938
\\
patio & 0.819 & 0.847 & 0.823 & 0.845 & 0.870
\\
wifi & 0.719 & 0.725 & 0.744 & 0.754 & 0.774
\\
\midrule
avg & 0.867 & 0.871 & 0.879 & 0.885 & 0.896
\\
\\
\end{tabular}
\caption{
Yelp reviews task:
Heldout AUC scores (higher is better) for various methods.
}%
\label{table:yelp_results}
\end{table}

%
\begin{table}
\begin{tabular}{l | c | r r r r}
\toprule
  prevalence
& drug
& Gibbs LDA
& PC LDA $\lambda = 100$
& BP sLDA
& logistic regr
\\
~
& ~
& K=200
& K=100
& K=100
&
\\
\midrule
  0.03
& nortriptyline
& 0.57	& 0.67	& 0.67	& 0.55
\\
  0.04
& mirtazapine
&0.66 &	0.69 &	0.71 &	0.58
\\
  0.05
&  escitalopram
& 0.57	& 0.59	& 0.62 & 0.61
\\
  0.05
& amitriptyline
& 0.67 &	0.71 &	0.73 &	0.64
\\
0.05 
& venlafaxine
& 0.60	& 0.63	& 0.59 &	0.61
\\
0.08
& paroxetine
& 0.62 &	0.66 &	0.61 &	0.61
\\
0.13
& bupropion
& 0.56	& 0.61 &	0.59 &	0.59
\\
0.16
& fluoxetine
& 	0.57 &	0.60 &	0.59 &	0.56
\\
0.16
& sertraline
& 0.58	& 0.59 &	0.61 &	0.58
\\
0.24 
& citalopram
&	0.59 &	0.60 &	0.58 &	0.59
\end{tabular}
\caption{
Antidepressant task:
Heldout AUC scores (higher is better) for various methods.
Each drug prediction task uses a common set of topic-word parameters but independent regression weights.
}%
\label{table:psych_results}
\end{table}

\paragraph{Prediction-constrained LDA's gains in predictive performance do not harm its heldout data predictions nearly as much as BP sLDA.}  The right panels of figures~\ref{fig:toy_letters_heldout_metrics}, ~\ref{fig:movie_review_heldout_metrics}, ~and~\ref{fig:yelp} show the performance of our PC-LDA approach and baselines on the generative task of modeling the data.  As expected, the Gibbs and Variational Bayes inference procedures applied to the unsupervised objective do best, because they are not trying to simultaneously optimize for any other task.  VB-sLDA also models $p(x)$ well, but as noted earlier, the lack of a weighted objective in the traditional sLDA formulation means that it also essentially focuses entirely on the generative task.  Of the two approaches that predict $p(y|x)$ well, BP-sLDA is consistently the worst performer in $p(x)$, with significantly more test error than PC-LDA.  
See for example the difference of well over 0.4 nats per token for $K=100$ topics on the Yelp dataset in Fig.~\ref{fig:yelp} between the gold line (BP-sLDA) and the dark red line (PC-LDA $\lambda = 1000$).
These results show that we can't expect a solely discriminative approach to explain the data well.
However, our prediction-constrained approach can use its model capacity wisely to capture the most variation in the data while getting high quality discriminative performance.

\paragraph{Prediction-constrained LDA topic-word parameters are qualitatively interpretable.} Fig.~\ref{fig:toy_letters_large} and Fig.~\ref{fig:psych_topics} show learned topics on the toy letters task and the antidepressant recommendation task.  On the letters task, we see that only PC-LDA and BP-sLDA achieve low error rates on the discriminative task; of those PC-LDA has features that look like letters---and in particular, vowels---while BP-sLDA's topics are less sparse and harder to interpret.  In the antidepressant recommendation task, there are ten drugs of interest.  Fig.~\ref{fig:psych_topics} shows the top medical codewords for the topics most predictive of success and non-success for the drug bupropion.  Again, the PC-LDA topics are clinically coherent: the top predictor of success contains words related to migraines, whereas the top predictor of non-success is concerned with testicular function.  In contrast, the topics found by BP-sLDA have little clinical coherence (as confirmed by a clinical collaborator).  Especially when the data are high-dimensional, having coherent topics from our dimensionality reduction---as well as high predictive performance---enables conversations with domain experts about what factors are most predictive of treatment success.

\begin{figure}
\begin{tabular}{c}
\verbatimfont{\tiny}%
\begin{minipage}{0.25\textwidth}
\begin{verbatim}
     BP sLDA + 7.7
0.60 nortriptyline 
0.27 nonspecific_abnormal_find 
0.21 other_specified_local_inf 
0.20 embryonic_cyst_of_fallopi 
0.20 supraspinatus_(muscle)_(t 
0.18 application_of_interverte 
0.16 other_malignant_neoplasm_ 
0.15 amoxicillin/clarithromyci 
0.15 need_for_prophylactic_vac 
0.15 observation_or_inpatient_
\end{verbatim}
\end{minipage}%
\begin{minipage}{0.25\textwidth}
\begin{verbatim}
     BP sLDA -15.8
0.39 visual_field_defect,_unsp 
0.39 citalopram 
0.36 microdissection_(ie,_samp 
0.35 need_for_prophylactic_vac 
0.31 pet_imaging_regional_or_w 
0.29 visual_discomfort 
0.29 accident_poison_by_heroin 
0.29 personal_history_of_alcoh
0.27 other_specified_intestina 
0.27 counseling_on_substance_u 
\end{verbatim}
\end{minipage}%
\begin{minipage}{0.25\textwidth}
\begin{verbatim}
     PCLDA + 3.8
0.99 migraine,_unspecified,_wi 
0.99 other_malaise_and_fatigue 
0.99 common_migraine,_without_ 
0.99 sumatriptan 
0.99 asa/butalbital/caffeine 
0.99 zolmitriptan 
0.99 migraine,_unspecified 
0.99 classical_migraine,_with_
0.99 classical_migraine,_witho 
0.99 migraine,_unspecified,_wi
\end{verbatim}
\end{minipage}%
\begin{minipage}{0.25\textwidth}
\begin{verbatim}
     PCLDA -26.4
1.00 semen_analysis;_complete_ 
1.00 male_infertility,_unspeci
1.00 lipoprotein,_direct_measu 
0.99 sperm_isolation;_simple_p 
0.99 tissue_culture_for_non-ne 
0.99 conditions_due_to_anomaly
0.99 vasectomy,_unilateral_or_ 
0.99 arthrocentesis 
0.99 scrotal_varices 
0.99 other_musculoskeletal_sym 
\end{verbatim}
\end{minipage}
\end{tabular}
\caption{-
Antidepressant task:
Visualization of top 2 (of 100) learned topics, selected by largest negative and positive logistic regression coefficients for the drug bupropion, for both BP sLDA (left) and our PCLDA $\lambda = 100$ (right).
PCLDA topics appear more interpretable and can guide conversations with clinicians about hypotheses to explore: e.g. ``are some drugs better for patients with history of migraines?''. In contrast, BP sLDA's exclusive focus on $p(y | x)$ makes its topics hard to interpret.
Each panel shows a topic's top medical codewords from the EHR ranked by $p( \mbox{topic} | \mbox{word} )$, computable via Bayes rule from the learn topic-word probabilities $\phi$.
}%
\label{fig:psych_topics}
\end{figure}

\section{Discussion and Conclusion}

Arriving at our proposed prediction-constrained training objective required many false starts. Below, we comment on two key advantages of our approach: designing our inference around modern gradient descent methods
and
focusing on asymmetry of the prediction task.
We also discuss two key limitations -- computational scalability and local optima -- and offer some possible remedies.

\paragraph{Building on modern gradient descent.}
We have designed our inference around modern stochastic gradient descent methods using automatic differentiation to compute the gradients.
This choice stands in contrast to prior work in supervised topic models using hand-designed coordinate descent \citep{blei2003lda},
mirror descent \citep{chen2015bplda}, or Gibbs sampling methods \citep{griffiths:2004:fst}. 
With automatic differentiation, it is very easy for practitioners to extend our efforts to custom loss functions for the supervised task without time-consuming derivations. For example, we were easily able to handle the multiple binary labels case for the Yelp reviews prediction task and antidepressant prediction task.

\paragraph{Focus on asymmetry.}
Our focus on asymmetry is new among the supervised topic model work we are aware of. 
More broadly, other authors, such as \citet{liu2009modularization,molitor2009using} describe asymmetric inference strategies (called ``cut distributions''), which result in principled probability distributions that are not the posterior of any graphical
model~\citep{plummer2015cuts}.

\paragraph{Towards better local optima.}
We found even with modern gradient methods, training our models via the PC objective was quite challenging: requiring many hours of computation and many random restarts to avoid local optima. For example, we see that even taking the best of 10 restarts, the PC-LDA $\lambda=100$ curve in Fig.~\ref{fig:yelp} shows poor performance at $K=100$ which we have isolated as a local optima (other methods' solutions are scored better by the PC-LDA objective). We expect some combination of better initialization procedures, annealing the objective, and intelligent proposal moves could lead to better local optima.

Our approach requires repeatedly solving the PC objective for different values of $\lambda$.  \citet{corduneanu2002continuation}
look at continuation or homotopy methods which balance multiple objectives via a tradeoff scalar parameter $\lambda \in [0, 1]$; starting from the unsupervised solution ($\lambda=0$) they gradually increase $\lambda$ and re-optimize. This approach was later applied by \citet{ji2009semisupervisedHomotopyHMM} to semi-supervised training of HMMs.  
Unfortunately, we found that the non-convexity of our objectives caused even small changes in $\lambda$ to induce solutions of the parameters $\xi$ that appear not to be connected to previous optima, so we do not recommend this as a practical way forward.

A similar concern about non-convexity occurs when using intelligent initialization strategies based on the purely unsupervised objective ($\lambda = 0$), such as topic model methods using 2nd or 3rd order word cooccurance moments, including ``anchor words'' \citep{arora2013anchor_topics}
or ``spectral methods''
\citep{ren2017spectral}.
More thorough comparison is needed, but our brief tests with using unsupervised methods as initializaitons suggest that the PC-LDA optimization landscape does not benefit from using these as initializations for gradient descent.  We found that parameter estimates often remain trapped near the unsupervised optima, unable to find parameters that produce better label predictions than more random initializations.

\paragraph{Towards more scalable training.}
Within our topic model case study, taking derivatives through the MAP embedding procedure is a significant runtime bottleneck.
One more scalable possibility would be to try to amortize this cost via a recognition network or variational auto-encoder (VAE) \citep{kingma2014autoencodingVB,srivastava2017autoencoding}. We briefly explored a recognition network for PC-training of supervised topic models in a previous workshop paper~\citep{hughes2016sLDARecogNetwork}, but we found the predictions of the VAE to be in general of lower quality than simply embedding the inference of the hidden variables within the gradient descent. We hope this report inspires future work so that our proposed PC objective can be easily applied to many LVMs.

\paragraph{Conclusion.}
We have presented a new optimization objective, the \emph{prediction-constrained} framework, for training latent variable models.
While previous methods are only appropriate for either fully discriminative or fully generative goals, 
our objective is unique in simultaneously \emph{balancing} these goals, allowing a practitioner to find the best possible generative model which meets some minimum prediction performance.
Our approach can also be applied in the semi-supervised setting, as we demonstrated in the mixtures case study. It is in this semi-supervised setting where we expect latent variable models to show the strongest advantages on prediction tasks.


%



\newpage
{\small
\linespread{1}
\bibliography{macros_for_journal_names,references}

\begin{thebibliography}{78}
\providecommand{\natexlab}[1]{#1}
\providecommand{\url}[1]{\texttt{#1}}
\expandafter\ifx\csname urlstyle\endcsname\relax
  \providecommand{\doi}[1]{doi: #1}\else
  \providecommand{\doi}{doi: \begingroup \urlstyle{rm}\Url}\fi

\bibitem[Abadi et~al.(2015)Abadi, Agarwal, Barham, Brevdo, Chen, Citro,
  Corrado, Davis, Dean, Devin, Ghemawat, Goodfellow, Harp, Irving, Isard, Jia,
  Jozefowicz, Kaiser, Kudlur, Levenberg, Man\'{e}, Monga, Moore, Murray, Olah,
  Schuster, Shlens, Steiner, Sutskever, Talwar, Tucker, Vanhoucke, Vasudevan,
  Vi\'{e}gas, Vinyals, Warden, Wattenberg, Wicke, Yu, and
  Zheng]{tensorflow2015whitepaper}
M.~Abadi, A.~Agarwal, P.~Barham, E.~Brevdo, Z.~Chen, C.~Citro, G.~S. Corrado,
  A.~Davis, J.~Dean, M.~Devin, S.~Ghemawat, I.~Goodfellow, A.~Harp, G.~Irving,
  M.~Isard, Y.~Jia, R.~Jozefowicz, L.~Kaiser, M.~Kudlur, J.~Levenberg,
  D.~Man\'{e}, R.~Monga, S.~Moore, D.~Murray, C.~Olah, M.~Schuster, J.~Shlens,
  B.~Steiner, I.~Sutskever, K.~Talwar, P.~Tucker, V.~Vanhoucke, V.~Vasudevan,
  F.~Vi\'{e}gas, O.~Vinyals, P.~Warden, M.~Wattenberg, M.~Wicke, Y.~Yu, and
  X.~Zheng.
\newblock {TensorFlow}: Large-scale machine learning on heterogeneous systems,
  2015.
\newblock URL \url{http://tensorflow.org/}.
\newblock Software available from tensorflow.org.

\bibitem[Al-Harbi and Rayward-Smith(2006)]{al2006adapting}
S.~H. Al-Harbi and V.~J. Rayward-Smith.
\newblock Adapting k-means for supervised clustering.
\newblock \emph{Applied Intelligence}, 24\penalty0 (3):\penalty0 219--226,
  2006.

\bibitem[Andrieu et~al.(2003)Andrieu, De~Freitas, Doucet, and
  Jordan]{andrieu2003introMCMC}
C.~Andrieu, N.~De~Freitas, A.~Doucet, and M.~I. Jordan.
\newblock An introduction to {MCMC} for machine learning.
\newblock \emph{Machine Learning}, 50\penalty0 (1-2):\penalty0 5--43, 2003.

\bibitem[Arora et~al.(2013)Arora, Ge, Halpern, Mimno, Moitra, Sontag, Wu, and
  Zhu]{arora2013anchor_topics}
S.~Arora, R.~Ge, Y.~Halpern, D.~Mimno, A.~Moitra, D.~Sontag, Y.~Wu, and M.~Zhu.
\newblock A practical algorithm for topic modeling with provable guarantees.
\newblock In \emph{International Conference on Machine Learning}, 2013.

\bibitem[Blei(2012)]{blei2012topicmodels}
D.~M. Blei.
\newblock Probabilistic topic models.
\newblock \emph{Communications of the ACM}, 55\penalty0 (4):\penalty0 77--84,
  2012.

\bibitem[Blei and Lafferty(2006)]{blei2006dynamic}
D.~M. Blei and J.~D. Lafferty.
\newblock Dynamic topic models.
\newblock In \emph{International Conference on Machine Learning}, 2006.

\bibitem[Blei et~al.(2003)Blei, Ng, and Jordan]{blei2003lda}
D.~M. Blei, A.~Y. Ng, and M.~I. Jordan.
\newblock Latent {D}irichlet allocation.
\newblock \emph{Journal of Machine Learning Research}, 3:\penalty0 993--1022,
  2003.

\bibitem[Chang et~al.(2007)Chang, Ratinov, and
  Roth]{chang2007semisupervizedwithconstraint}
M.-W. Chang, L.~Ratinov, and D.~Roth.
\newblock Guiding semi-supervision with constraint-driven learning.
\newblock In \emph{Proc. of the Annual Meeting of the Association for
  Computational Linguistics}, 2007.

\bibitem[Chen et~al.(2015)Chen, He, Shen, Xiao, He, Gao, Song, and
  Deng]{chen2015bplda}
J.~Chen, J.~He, Y.~Shen, L.~Xiao, X.~He, J.~Gao, X.~Song, and L.~Deng.
\newblock End-to-end learning of {LDA} by mirror-descent back propagation over
  a deep architecture.
\newblock In \emph{Neural Information Processing Systems}, 2015.

\bibitem[Corduneanu and Jaakkola(2002)]{corduneanu2002continuation}
A.~Corduneanu and T.~Jaakkola.
\newblock Continuation methods for mixing heterogeneous sources.
\newblock In \emph{Uncertainty in Artificial Intelligence}, 2002.

\bibitem[Crouse et~al.(1998)Crouse, Nowak, and Baraniuk]{crouse1998hmt}
M.~S. Crouse, R.~D. Nowak, and R.~G. Baraniuk.
\newblock Wavelet-based statistical signal processing using hidden {M}arkov
  models.
\newblock \emph{IEEE Transactions on Signal Processing}, 46\penalty0
  (4):\penalty0 886--902, 1998.

\bibitem[Dhurandhar et~al.(2017)Dhurandhar, Ackerman, and
  Wang]{dhurandhar2017uncovering}
A.~Dhurandhar, M.~Ackerman, and X.~Wang.
\newblock Uncovering group level insights with accordant clustering.
\newblock \emph{arXiv preprint arXiv:1704.02378}, 2017.

\bibitem[DiCicco and Patel(2010)]{dicicco2010machine}
T.~M. DiCicco and R.~Patel.
\newblock Machine classification of prosodic control in dysarthria.
\newblock \emph{Journal of medical speech-language pathology}, 18\penalty0
  (4):\penalty0 35, 2010.

\bibitem[Eick et~al.(2004)Eick, Zeidat, and Zhao]{eick2004supervised}
C.~F. Eick, N.~Zeidat, and Z.~Zhao.
\newblock Supervised clustering-algorithms and benefits.
\newblock In \emph{Tools with Artificial Intelligence, 2004. ICTAI 2004. 16th
  IEEE International Conference on}, pages 774--776. IEEE, 2004.

\bibitem[Everitt and Hand(1981)]{everitt1981mixtures}
B.~S. Everitt and D.~Hand.
\newblock \emph{Finite mixture distributions}.
\newblock Chapman and Hall, 1981.
\newblock ISBN 0412224208.

\bibitem[Finley and Joachims(2005)]{finley2005supervised}
T.~Finley and T.~Joachims.
\newblock Supervised clustering with support vector machines.
\newblock In \emph{Proceedings of the 22nd international conference on Machine
  learning}, pages 217--224. ACM, 2005.

\bibitem[Flammarion et~al.(2016)Flammarion, Palaniappan, and
  Bach]{flammarion2016robust}
N.~Flammarion, B.~Palaniappan, and F.~Bach.
\newblock Robust discriminative clustering with sparse regularizers.
\newblock \emph{arXiv preprint arXiv:1608.08052}, 2016.

\bibitem[Ganchev et~al.(2010)Ganchev, Gra\c{c}a, Gillenwater, and
  Taskar]{ganchev2010posteriorconstraints}
K.~Ganchev, J.~Gra\c{c}a, J.~Gillenwater, and B.~Taskar.
\newblock Posterior regularization for structured latent variable models.
\newblock \emph{Journal of Machine Learning Research}, 11:\penalty0 2001--2049,
  Aug. 2010.

\bibitem[Ghahramani and Hinton(1996)]{ghahramani1996estimationForLDS}
Z.~Ghahramani and G.~E. Hinton.
\newblock Parameter estimation for linear dynamical systems.
\newblock Technical Report CRG-TR-96-2, University of Toronto Dept. of Computer
  Science, 1996.

\bibitem[Ghahramani and Jordan(1993)]{ghahramani1993supervisedEMforMixtures}
Z.~Ghahramani and M.~I. Jordan.
\newblock Supervised learning from incomplete data via an em approach.
\newblock In \emph{Neural Information Processing Systems}, 1993.

\bibitem[Gra\c{c}a et~al.(2008)Gra\c{c}a, Ganchev, and
  Taskar]{gracca2008posteriorconstraints}
J.~Gra\c{c}a, K.~Ganchev, and B.~Taskar.
\newblock Expectation maximization and posterior constraints.
\newblock In \emph{Neural Information Processing Systems}, 2008.

\bibitem[Grbovic et~al.(2013)Grbovic, Djuric, Guo, and
  Vucetic]{grbovic2013supervised}
M.~Grbovic, N.~Djuric, S.~Guo, and S.~Vucetic.
\newblock Supervised clustering of label ranking data using label preference
  information.
\newblock \emph{Machine learning}, 93\penalty0 (2-3):\penalty0 191--225, 2013.

\bibitem[Griffiths and Ghahramani(2007)]{griffiths2007ibp}
T.~L. Griffiths and Z.~Ghahramani.
\newblock Infinite latent feature models and the {I}ndian buffet process.
\newblock In \emph{Neural Information Processing Systems}, 2007.

\bibitem[Griffiths and Steyvers(2004)]{griffiths:2004:fst}
T.~L. Griffiths and M.~Steyvers.
\newblock Finding scientific topics.
\newblock \emph{Proceedings of the National Academy of Sciences}, 2004.

\bibitem[Halpern et~al.(2012)Halpern, Horng, Nathanson, Shapiro, and
  Sontag]{halpern2012comparison}
Y.~Halpern, S.~Horng, L.~A. Nathanson, N.~I. Shapiro, and D.~Sontag.
\newblock A comparison of dimensionality reduction techniques for unstructured
  clinical text.
\newblock In \emph{{ICML} workshop on clinical data analysis}, 2012.

\bibitem[Hannah et~al.(2011)Hannah, Blei, and Powell]{hannah2011DPmixGLM}
L.~A. Hannah, D.~M. Blei, and W.~B. Powell.
\newblock Dirichlet process mixtures of generalized linear models.
\newblock \emph{Journal of Machine Learning Research}, 12\penalty0
  (Jun):\penalty0 1923--1953, 2011.

\bibitem[Hughes et~al.(2016)Hughes, Elibol, McCoy, Perlis, and
  Doshi-Velez]{hughes2016sLDARecogNetwork}
M.~C. Hughes, H.~M. Elibol, T.~McCoy, R.~Perlis, and F.~Doshi-Velez.
\newblock Supervised topic models for clinical interpretability.
\newblock \emph{arXiv preprint arXiv:1612.01678v1}, 2016.

\bibitem[Ismaili et~al.(2016)Ismaili, Lemaire, and
  Cornu{\'e}jols]{ismaili2016supervised}
O.~A. Ismaili, V.~Lemaire, and A.~Cornu{\'e}jols.
\newblock Supervised pre-processings are useful for supervised clustering.
\newblock In \emph{Analysis of Large and Complex Data}, pages 147--157.
  Springer, 2016.

\bibitem[Jaakkola et~al.(1999{\natexlab{a}})Jaakkola, Meila, and
  Jebara]{jaakkola1999MEDtechreport}
T.~S. Jaakkola, M.~Meila, and T.~Jebara.
\newblock Maximum entropy discrimination.
\newblock Technical Report AITR-1668, Artificial Intelligence Laboratory at the
  Massachusetts Institute of Technology, 1999{\natexlab{a}}.
\newblock URL \url{http://people.csail.mit.edu/tommi/papers/maxent.ps}.

\bibitem[Jaakkola et~al.(1999{\natexlab{b}})Jaakkola, Meila, and
  Jebara]{jaakkola1999med}
T.~S. Jaakkola, M.~Meila, and T.~Jebara.
\newblock Maximum entropy discrimination.
\newblock In \emph{Neural Information Processing Systems}, 1999{\natexlab{b}}.

\bibitem[Jebara(2001)]{jebara2001medthesis}
T.~Jebara.
\newblock \emph{Discriminative, generative and imitative learning}.
\newblock PhD thesis, Massachusetts Institute of Technology, 2001.

\bibitem[Jebara and Pentland(1999)]{jebara1999cem}
T.~Jebara and A.~Pentland.
\newblock Maximum conditional likelihood via bound maximization and the {CEM}
  algorithm.
\newblock In \emph{Neural Information Processing Systems}, 1999.

\bibitem[Ji et~al.(2009)Ji, Watson, and Carin]{ji2009semisupervisedHomotopyHMM}
S.~Ji, L.~T. Watson, and L.~Carin.
\newblock Semisupervised learning of hidden markov models via a homotopy
  method.
\newblock \emph{IEEE Transactions on Pattern Analysis and Machine
  Intelligence}, 31\penalty0 (2):\penalty0 275--287, 2009.

\bibitem[Jiang et~al.(2015)Jiang, Qian, Shen, and Mei]{jiang2015travel}
S.~Jiang, X.~Qian, J.~Shen, and T.~Mei.
\newblock Travel recommendation via author topic model based collaborative
  filtering.
\newblock In \emph{International Conference on Multimedia Modeling}, pages
  392--402. Springer, 2015.

\bibitem[Kemp et~al.(2006)Kemp, Tenenbaum, Griffiths, Yamada, and
  Ueda]{kemp2006irm}
C.~Kemp, J.~B. Tenenbaum, T.~L. Griffiths, T.~Yamada, and N.~Ueda.
\newblock Learning systems of concepts with an infinite relational model.
\newblock In \emph{AAAI Conference on Artificial Intelligence}, 2006.

\bibitem[Kingma and Ba(2014)]{kingma2014adam}
D.~Kingma and J.~Ba.
\newblock Adam: A method for stochastic optimization.
\newblock \emph{arXiv preprint arXiv:1412.6980}, 2014.

\bibitem[Kingma and Welling(2014)]{kingma2014autoencodingVB}
D.~Kingma and M.~Welling.
\newblock Auto-encoding variational {B}ayes.
\newblock In \emph{The International Conference on Learning Representations
  (ICLR)}, 2014.

\bibitem[Kivinen and Warmuth(1997)]{kivinen1997exponentiated_gradient}
J.~Kivinen and M.~K. Warmuth.
\newblock Exponentiated gradient versus gradient descent for linear predictors.
\newblock \emph{Information and Computation}, 132\penalty0 (1):\penalty0 1--63,
  1997.

\bibitem[Krause et~al.(2010)Krause, Perona, and
  Gomes]{krause2010discriminative}
A.~Krause, P.~Perona, and R.~G. Gomes.
\newblock Discriminative clustering by regularized information maximization.
\newblock In \emph{Advances in neural information processing systems}, pages
  775--783, 2010.

\bibitem[Lacoste-Julien et~al.(2009)Lacoste-Julien, Sha, and
  Jordan]{lacoste2009disclda}
S.~Lacoste-Julien, F.~Sha, and M.~I. Jordan.
\newblock Disc{L}{D}{A}: Discriminative learning for dimensionality reduction
  and classification.
\newblock In \emph{Neural Information Processing Systems}, 2009.

\bibitem[Liu et~al.(2009)Liu, Bayarri, Berger, et~al.]{liu2009modularization}
F.~Liu, M.~Bayarri, J.~Berger, et~al.
\newblock Modularization in bayesian analysis, with emphasis on analysis of
  computer models.
\newblock \emph{Bayesian Analysis}, 4\penalty0 (1):\penalty0 119--150, 2009.

\bibitem[Liverani et~al.(2015)Liverani, Hastie, Azizi, Papathomas, and
  Richardson]{liverani2015premium}
S.~Liverani, D.~Hastie, L.~Azizi, M.~Papathomas, and S.~Richardson.
\newblock {PReMiuM}: An {R} package for profile regression mixture models using
  {D}irichlet processes.
\newblock \emph{Journal of Statistical Software}, 64\penalty0 (7):\penalty0
  1--30, 2015.
\newblock URL \url{https://www.jstatsoft.org/v064/i07}.

\bibitem[MacKay(1997)]{mackay1997ensemble}
D.~J.~C. MacKay.
\newblock Ensemble learning for hidden {M}arkov models.
\newblock Technical report, Department of Physics, University of Cambridge,
  1997.

\bibitem[Maclaurin et~al.(2015)Maclaurin, Duvenaud, Johnson, and
  Adams]{maclaurin2015autograd}
D.~Maclaurin, D.~Duvenaud, M.~Johnson, and R.~Adams.
\newblock Autograd: Reverse-mode differentiation of native python.
\newblock \url{http://github. com/HIPS/autograd}, 2015.

\bibitem[Mann and McCallum(2007)]{mann2007expectationRegularization}
G.~S. Mann and A.~McCallum.
\newblock Simple, robust, scalable semi-supervised learning via expectation
  regularization.
\newblock In \emph{International Conference on Machine Learning}, 2007.

\bibitem[Mann and McCallum(2010)]{mann2010generalized}
G.~S. Mann and A.~McCallum.
\newblock Generalized expectation criteria for semi-supervised learning with
  weakly labeled data.
\newblock \emph{Journal of Machine Learning Research}, 11\penalty0
  (Feb):\penalty0 955--984, 2010.

\bibitem[Mc{A}uliffe and Blei(2007)]{blei2007sLDA}
J.~D. Mc{A}uliffe and D.~M. Blei.
\newblock Supervised topic models.
\newblock In \emph{Neural Information Processing Systems}, 2007.

\bibitem[McCallum(2002)]{MALLET}
A.~K. McCallum.
\newblock {MALLET}: Machine learning for language toolkit.
\newblock \url{mallet.cs.umass.edu}, 2002.

\bibitem[Mimno and Mc{C}allum(2008)]{mimno2008dmr}
D.~Mimno and A.~Mc{C}allum.
\newblock Topic models conditioned on arbitrary features with
  {D}irichlet-multinomial regression.
\newblock In \emph{Uncertainty in Artificial Intelligence}, 2008.

\bibitem[Molitor et~al.(2010)Molitor, Papathomas, Jerrett, and
  Richardson]{molitor2010bayesianProfileRegression}
J.~Molitor, M.~Papathomas, M.~Jerrett, and S.~Richardson.
\newblock Bayesian profile regression with an application to the national
  survey of children's health.
\newblock \emph{Biostatistics}, 11\penalty0 (3):\penalty0 484--498, 2010.

\bibitem[Molitor et~al.(2009)Molitor, Best, Jackson, and
  Richardson]{molitor2009using}
N.-T. Molitor, N.~Best, C.~Jackson, and S.~Richardson.
\newblock Using bayesian graphical models to model biases in observational
  studies and to combine multiple sources of data: application to low birth
  weight and water disinfection by-products.
\newblock \emph{Journal of the Royal Statistical Society: Series A (Statistics
  in Society)}, 172\penalty0 (3):\penalty0 615--637, 2009.

\bibitem[Nigam et~al.(1998)Nigam, McCallum, Thrun, and
  Mitchell]{nigam1998semisuperEM}
K.~Nigam, A.~McCallum, S.~Thrun, and T.~Mitchell.
\newblock Learning to classify text from labeled and unla- beled documents.
\newblock In \emph{AAAI Conference on Artificial Intelligence}, 1998.

\bibitem[Pang and Lee(2005)]{pang2005moviereviews}
B.~Pang and L.~Lee.
\newblock Seeing stars: Exploiting class relationships for sentiment
  categorization with respect to rating scales.
\newblock In \emph{Proc. of the Annual Meeting of the Association for
  Computational Linguistics}, 2005.

\bibitem[Paul and Dredze(2012)]{paul2012model}
M.~J. Paul and M.~Dredze.
\newblock A model for mining public health topics from twitter.
\newblock \emph{Health}, 11:\penalty0 16--6, 2012.

\bibitem[Peralta et~al.(2013)Peralta, Espinace, and Soto]{peralta2013enhancing}
B.~Peralta, P.~Espinace, and A.~Soto.
\newblock Enhancing k-means using class labels.
\newblock \emph{Intelligent Data Analysis}, 17\penalty0 (6):\penalty0
  1023--1039, 2013.

\bibitem[Peralta et~al.(2016)Peralta, Caro, and Soto]{peralta2016proposal}
B.~Peralta, A.~Caro, and A.~Soto.
\newblock A proposal for supervised clustering with dirichlet process using
  labels.
\newblock \emph{Pattern Recognition Letters}, 80:\penalty0 52--57, 2016.

\bibitem[Plummer(2015)]{plummer2015cuts}
M.~Plummer.
\newblock Cuts in {B}ayesian graphical models.
\newblock \emph{Statistics and Computing}, 25\penalty0 (1):\penalty0 37--43,
  2015.

\bibitem[Rabiner and Juang(1986)]{rabiner1986introduction}
L.~R. Rabiner and B.-H. Juang.
\newblock An introduction to hidden markov models.
\newblock \emph{ASSP Magazine, IEEE}, 3\penalty0 (1):\penalty0 4--16, 1986.

\bibitem[Ramage et~al.(2009)Ramage, Hall, Nallapati, and
  Manning]{ramage2009labeled}
D.~Ramage, D.~Hall, R.~Nallapati, and C.~D. Manning.
\newblock Labeled {LDA}: A supervised topic model for credit attribution in
  multi-labeled corpora.
\newblock In \emph{Proceedings of the 2009 Conference on Empirical Methods in
  Natural Language Processing: Volume 1-Volume 1}, pages 248--256. Association
  for Computational Linguistics, 2009.

\bibitem[Ramani and Jacob(2013)]{ramani2013improved}
R.~G. Ramani and S.~G. Jacob.
\newblock Improved classification of lung cancer tumors based on structural and
  physicochemical properties of proteins using data mining models.
\newblock \emph{PloS one}, 8\penalty0 (3):\penalty0 e58772, 2013.

\bibitem[Ren et~al.(2017)Ren, Wang, and Zhu]{ren2017spectral}
Y.~Ren, Y.~Wang, and J.~Zhu.
\newblock Spectral learning for supervised topic models.
\newblock \emph{IEEE Transactions on Pattern Analysis and Machine
  Intelligence}, 2017.

\bibitem[Roweis(1998)]{roweis1998algorithms}
S.~T. Roweis.
\newblock {EM} algorithms for {PCA} and {SPCA}.
\newblock In \emph{Neural Information Processing Systems}, 1998.

\bibitem[Shahbaba and Neal(2009)]{shahbaba2009nonlinearDPmix}
B.~Shahbaba and R.~Neal.
\newblock Nonlinear models using dirichlet process mixtures.
\newblock \emph{Journal of Machine Learning Research}, 10\penalty0
  (Aug):\penalty0 1829--1850, 2009.

\bibitem[Shumway and Stoffer(1982)]{shumway1982emForLDS}
R.~H. Shumway and D.~S. Stoffer.
\newblock An approach to time series smoothing and forecasting using the {EM}
  algorithm.
\newblock \emph{Journal of {T}ime {S}eries {A}nalysis}, 3\penalty0
  (4):\penalty0 253--264, 1982.

\bibitem[Sontag and Roy(2011)]{sontag2011complexityoflda}
D.~Sontag and D.~Roy.
\newblock Complexity of inference in latent dirichlet allocation.
\newblock In \emph{Neural Information Processing Systems}, 2011.

\bibitem[Srivastava and Sutton(2017)]{srivastava2017autoencoding}
A.~Srivastava and C.~Sutton.
\newblock Autoencoding variational inference for topic models.
\newblock \emph{ICLR}, 2017.

\bibitem[Taddy(2012)]{taddy2012topicmodelmapestimation}
M.~Taddy.
\newblock On estimation and selection for topic models.
\newblock In \emph{Artificial Intelligence and Statistics}, 2012.

\bibitem[Tipping and Bishop(1999)]{tipping1999probabilisticPCA}
M.~E. Tipping and C.~M. Bishop.
\newblock Probabilistic principal component analysis.
\newblock \emph{Journal of the Royal Statistical Society: Series B (Statistical
  Methodology)}, 61\penalty0 (3):\penalty0 611--622, 1999.

\bibitem[Wainwright and Jordan(2008)]{wainwright2008variational}
M.~J. Wainwright and M.~I. Jordan.
\newblock Graphical models, exponential families, and variational inference.
\newblock \emph{Foundations and Trends{\textregistered} in Machine Learning},
  1\penalty0 (1-2):\penalty0 1--305, 2008.

\bibitem[Wang et~al.(2009)Wang, Blei, and Li]{wang2009simultaneous}
C.~Wang, D.~Blei, and F.-F. Li.
\newblock Simultaneous image classification and annotation.
\newblock In \emph{IEEE Conf. on Computer Vision and Pattern Recognition},
  2009.

\bibitem[Wang and Zhu(2014)]{wang2014spectral}
Y.~Wang and J.~Zhu.
\newblock Spectral methods for supervised topic models.
\newblock In \emph{Advances in Neural Information Processing Systems}, pages
  1511--1519, 2014.

\bibitem[Wang and Wong(1987)]{wang1987stochasticBlockmodels}
Y.~J. Wang and G.~Y. Wong.
\newblock Stochastic blockmodels for directed graphs.
\newblock \emph{Journal of the American Statistical Association}, 82\penalty0
  (397):\penalty0 8--19, 1987.

\bibitem[{Y}elp Dataset~Challenge(2016)]{yelpdataset2016}
{Y}elp Dataset~Challenge.
\newblock {Y}elp dataset challenge.
\newblock \url{https://www.yelp.com/dataset_challenge}, 2016.
\newblock Accessed: 2016-03.

\bibitem[Yoon et~al.(2016)Yoon, Alaa, Cadeiras, and van~der
  Schaar]{yoon2016personalized}
J.~Yoon, A.~M. Alaa, M.~Cadeiras, and M.~van~der Schaar.
\newblock Personalized donor-recipient matching for organ transplantation.
\newblock \emph{arXiv preprint arXiv:1611.03934}, 2016.

\bibitem[Zhang and Kjellström(2014)]{zhang2014howToSuperviseTopicModels}
C.~Zhang and H.~Kjellström.
\newblock How to supervise topic models.
\newblock In \emph{ECCV Workshop on Graphical Models in Computer Vision}, 2014.

\bibitem[Zhu et~al.(2012)Zhu, Ahmed, and Xing]{zhu2012medlda}
J.~Zhu, A.~Ahmed, and E.~P. Xing.
\newblock Med{L}{D}{A}: maximum margin supervised topic models.
\newblock \emph{The Journal of Machine Learning Research}, 13\penalty0
  (1):\penalty0 2237--2278, 2012.

\bibitem[Zhu et~al.(2013)Zhu, Chen, Perkins, and Zhang]{zhu2013gibbsmaxmargin}
J.~Zhu, N.~Chen, H.~Perkins, and B.~Zhang.
\newblock Gibbs max-margin topic models with fast sampling algorithms.
\newblock In \emph{International Conference on Machine Learning}, 2013.

\bibitem[Zhu et~al.(2014)Zhu, Chen, and Xing]{zhu2014regbayes}
J.~Zhu, N.~Chen, and E.~P. Xing.
\newblock Bayesian inference with posterior regularization and applications to
  infinite latent svms.
\newblock \emph{Journal of Machine Learning Research}, 15\penalty0
  (1):\penalty0 1799--1847, 2014.

\end{thebibliography}
}

\begin{appendix}
\section{Datasets descriptions}

\subsection{Toy Vowels-from-consonants}

This dataset, each document is a count vector over a vocabulary where each symbol is a pixel in a 26 x 26 square grid. Data is generated from an LDA model with 30 total topics: 26 ``letter'' topics, each one displaying either a vowel (A, E, I, O, or U) or a consonant, as well as 4 ``background'' topics that have more dispersed distributions.
Each letter's active pixels are confined to a small region of the entire square grid, and thus several letters are easily confused: for example "E" and "F" share many of the same pixels, as do "I" and "J" and "O" and "M".
Each document has at least 2 and up to 4 topics that are active. Documents are labeled $y_d=1$ only if at least one vowel appears. 

The final corpus includes 10000 training documents, 500 validation documents, and 500 test-set documents. Each document has between 100 and 400 tokens.


\subsection{Movie reviews}
Raw text from movie reviews of four critics comes from scaledata v1.0 dataset released by Pang et al \citep{pang2005moviereviews}\footnote{\url{http://www.cs.cornell.edu/people/pabo/movie-review-data/}}.
Given plain text files of movie reviews, we tokenized and then stemmed using the Snowball stemmer from the nltk Python package, so that words with similar roots (e.g. film, films, filming) all become the same token. We removed all tokens in Mallet's list of common English stop words as well as any token included in the 1000 most common first names from the US census. We added this step after seeing too many common first names like Michael and Jennifer appear meaninglessly in many top-word lists for trained topics. We manually whitelisted "oscar" and "tony" due to their saliency to movie reviews sentiment. We then performed counts of all remaining tokens across the full raw corpus of 5006 documents, discarding any tokens that appear at least once in more than 20\% of all documents or less than 30 distinct documents. The final vocabulary list has 5375 terms.

Each of the 5006 original documents was then reduced to this vocabulary set. We discarded any documents that were too short (less than 20 tokens), leaving 5005 documents. Each document has a binary label, where 0 indicates it has a negative review (below 0.6 in the original datasets' 0-1 scale) and 1 indicates positive review (>= 0.6). This 0.6 threshold matches a threshold previously used in the raw data's 4-category scale to separate 0 and 1 star reviews from 2 and 3 (of 3) star reviews. Data pairs ($x_d, y_d$) were then split into training, validation, test. Both validation and test used 10 \% of all documents, evenly balancing positive and negative labeled documents. The remaining documents were allocated to the training set.


\subsection{Yelp reviews}

We use raw text of online Yelp reviews from the Yelp dataset challenge \citep{yelpdataset2016} to construct a multi-label binary dataset. This dataset includes text reviews about businesses. The businesses have associated meta data. We consider only businesses who have values for seven interesting binary attributes: “reservations accepted”,
“deliver offered”, “alcohol served”, “good for kids”, “price range > 1”
\footnote{
price range is given as an integer 1-4 where 1 is very cheap and 4 is very expensive. We turn this into a binary attribute by separating price range 1 from higher price ranges 2, 3 and 4.}, “outdoor seating” and “wifi offered”. To construct the documents, we concatenate all reviews about a single business. Thus, each business is represented by a single document. We also prune the vocabulary, removing rare words that occur in fewer than 5 documents and removing very common words that occur in more than 50\% of the documents. Finally, we sort the remaining words by tf-idf score and keep the top 10,000 scoring words as our final vocabulary. The resulting corpus includes 23,159 documents and a total of 43,236,060 observations.


Here is a summary of how often each binary label is positive in the dataset:
\begin{itemize}
\item reservations (41.9\% True)
\item delivery (20.7\% True)
\item alcohol (55.1\% True)
\item kid-friendly (83.4\% True)
\item expensive (61.0\% True)
\item patio (41.5\% True)
\item wifi (42.7\% True)
\end{itemize}

\subsection{Psychiatric EHR dataset}

We study a deidentified cohort of hundreds of thousands patients drawn from two large academic medical centers and their affiliated outpatient networks over a period of several years.
Each patient has at least one ICD9 diagnostic code for major depressive
disorder (ICD9s 296.2x or 3x or 311, or ICD10 equivalent).
Each included patient had an identified successful treatment
which included one of 25 possible common anti-depressants marked as ``primary'' treatments for major depressive disorder by clinical collaborators. 
We labeled an interval of a patient's record ``successful'' if 
all prescription events in the interval used the same subset of primary drugs, the interval lasted at least 90 days, and encounters occurred at least every 13 months. 
Applying this criteria, we identified 64431 patients who met our definition of success. For each patient, we extracted a bag-of-codewords $x_d$ of 7291 possible codewords (representing medical history before any successful treatment) and binary label vector $y_d$, marking which of 10 prevalent anti-depressants (if any) were used in known successful treatment.  The IRBs of Harvard University and Massachusetts General Hospital approved this study.

\paragraph{Extracting data $x_d$.}
For each patient with known successful treatment, we build a data vector $x_d$ to summarize all facts known about the patient in the EHR before any successful treatment was given.
Thus, we must confine our records to the interval from the patient's first encounter to the last encounter before any of the drugs on his or her successful list were first prescribed.
To summarize this patient's interval of ``pre-successful treatment'', 
we built a sparse count vector of all procedures, diagnoses, labs, and medications from the EHR which fit within the interval (22,000 possible codewords).
By definition, none of the anti-depressant medications on the patient's eventual success list appear in $x_d$.
To simplify, we reduced this to a final vocabulary of 7291 codewords that occurred in at least 1000 distinct patients.
We discard any patients with fewer than 5 tokens in $x_d$ (little to no history).

\paragraph{Extracting labels $y_d$.}
Among the 25 primary drugs, we identified a smaller set of 10 anti-depressants which were used in ``successful treatment'' for at least 1000 patients. The remaining 15 primary drugs did not occur commonly enough that we thought we could accurately access prediction quality. Our chosen list of drugs to predict are:
\begin{verbatim}
nortriptyline
amitriptyline
bupropion
fluoxetine
sertraline
paroxetine
venlafaxine
mirtazapine
citalopram
escitalopram
\end{verbatim}
Because these drugs can be given in combination, this is a multiple binary label problem. Future work could look into structured prediction tasks.

\paragraph{Public release:} Unfortunately, due to privacy concerns this dataset cannot be made public. Please contact the first author with questions.

\section{Protocol details for Topic Modeling Experiments}
Methods were allowed to run for 5000 complete passes through the dataset, or up to 48 hours.

\subsection{Step sizes}
PCLDA requires the choice of step size for Adam optimizer. We select among 0.01, 0.0333, and 0.1 using validation set. Generally, larger rates like 0.1 are preferred.  BP-sLDA also requires a step size, so we choose among 0.050,  0.010, 0.005.

\subsection{Initialization}
We consider two possible ways to initialize topics $\phi$. First, drawing all topics from low-variance random noise so no initial topic is too extreme yet symmetry breaking occurs. Second, using the anchor words procedure of \citep{arora2013anchor_topics}, which given $K$ finds a set of $K$ vocabulary words whose empirical distributions nicely cover the ``span'' of all observed word-cooccurance distributions.

\subsection{Multiple restarts}
For each possible hyperparameter setting, all methods were allowed 3 separate randomly-seeded initializations of topics $\phi$ and regression weights $\eta$. (We plan to run more but ran out of compute resources).

\subsection{Batch sizes}
We used different batch sizes for different data sets, as they were of different sizes and analysed on different computing architectures with different capabilities: 
\begin{itemize}
\item Toy letters dataset: 5 batches (2000 docs / batch)
\item Movie reviews: 1 batch (4004 docs / batch).
\item Yelp: 20 batches.
\item Psychiatric EHR: 20 batches.
\end{itemize}

\end{appendix}

\end{document}